\documentclass{article}

\usepackage{iclr2027_conference,times}

\usepackage[T1]{fontenc}
\usepackage[utf8]{inputenc}
\usepackage{microtype}
\usepackage{inconsolata}
\usepackage{xspace}
\usepackage{latexsym}
\usepackage{graphicx}
\usepackage{wrapfig}
\usepackage{capt-of}
\usepackage{float}
\usepackage{booktabs}
\usepackage{tabularx}
\usepackage{hyperref}
\usepackage{url}
\usepackage{multirow}
\usepackage{amsmath}
\usepackage{amssymb}
\usepackage[ruled,vlined]{algorithm2e}
\usepackage{mathtools}
\usepackage{bm}
\usepackage{xcolor}
\usepackage{colortbl}
\usepackage{tcolorbox}

\newcommand{\ours}{IVE\xspace}

\title{Attention at Rest Stays at Rest: \\ Breaking Visual Inertia to Mitigate Relation Hallucinations}

\author{
  Boyang Gong \\
  Tsinghua University \\
  \texttt{gongby25@mails.tsinghua.edu.cn}
  \And
  Yu Zheng$^{\dagger}$ \\
  Tsinghua University \\
  \texttt{yu-zheng@tsinghua.edu.cn}
  \And
  Fanye Kong \\
  Tsinghua University \\
  \texttt{kongfy23@mails.tsinghua.edu.cn}
  \AND
  Jie Zhou \\
  Tsinghua University \\
  \texttt{jzhou@tsinghua.edu.cn}
  \And
  Jiwen Lu \\
  Tsinghua University \\
  \texttt{lujiwen@tsinghua.edu.cn}
}

\iclrfinalcopy
\begin{document}

\maketitle
\lhead{}

\begingroup
\setlength{\intextsep}{2pt}
\setlength{\abovecaptionskip}{2pt}
\vspace{-0.16in}
\begin{figure}[H]
    \centering
    \includegraphics[width=\linewidth]{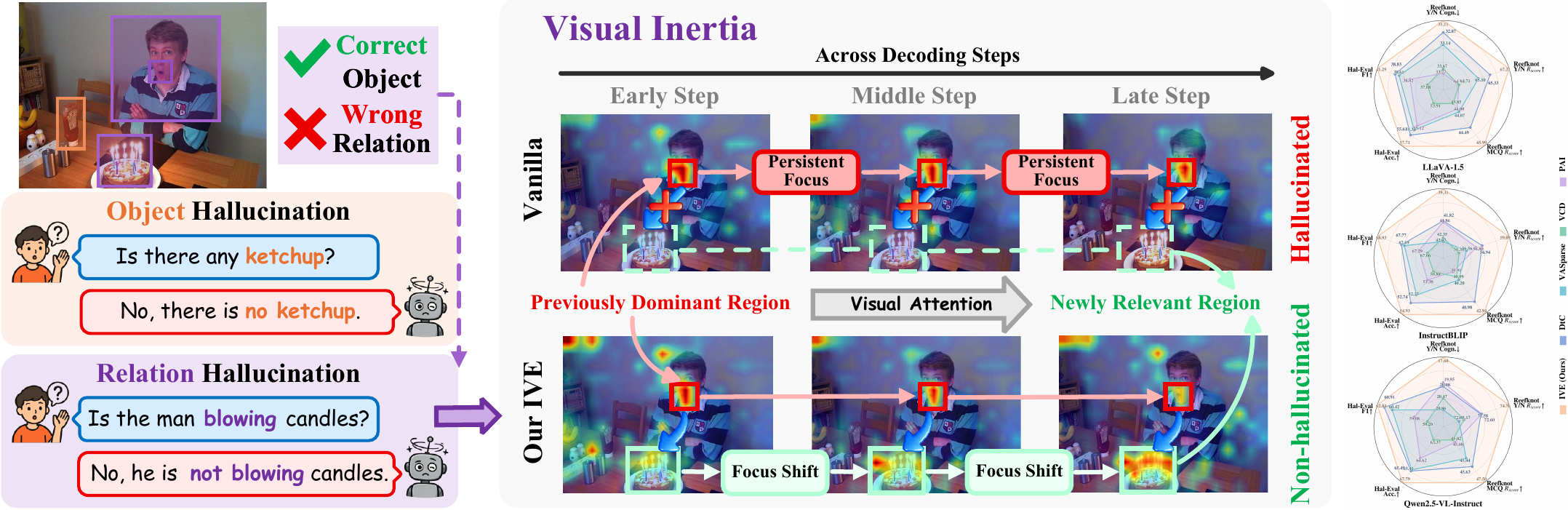}
    \caption{
    \textbf{Overview of Visual Inertia in relation hallucinations.}
    \textbf{Left:} Object hallucination concerns entity presence, whereas relation hallucination can occur despite correct object recognition.
    \textbf{Middle:} Visual attention remains focused on previously dominant regions across decoding rather than shifting toward newly relevant evidence, illustrating visual inertia.
    \textbf{Right:} Radar comparison of \ours and competing methods across three MLLM backbones on Reefknot and Hal-Eval.
    }
    \label{fig:teaser}
\end{figure}
\endgroup

\begin{abstract}
While multimodal large language models demonstrate strong entity-level perception, faithfully grounding relational interactions between objects remains a persistent challenge. 
Although conventional visual grounding techniques attempt to resolve hallucinations by amplifying visual attention, strengthening overall visual signals fails to reliably correct relational errors. 
Tracing visual attention in relation descriptions reveals that correct responses tend to dynamically shift focus across regions, whereas hallucinated responses often linger on previously dominant evidence, exhibiting an undesirable \textit{visual inertia}.
Further analysis shows that relation-prediction performance steadily deteriorates as more previous-step visual attention is carried into the current decoding step.
We therefore introduce Inertia-aware Visual Excitation (\ours), an MLLM decoding method that dynamically recalibrates visual values using token-level attention history.
By contrasting current attention against recent moving averages, IVE separates emergent tokens with rising relevance from persistently dominant inertia tokens, selectively reinforcing newly needed evidence while mildly attenuating contributions from repeatedly attended regions. 
Across three MLLMs and decoding strategies, \ours reduces relation hallucinations while preserving broader multimodal performance.
\end{abstract}

\section{Introduction}
\label{sec:intro}

Multimodal large language models (MLLMs)~\cite{liu2024visual,liu2024improved,li2023blip,zhu2023minigpt,bai2025qwen2} combine visual and textual information to generate natural-language responses. Object-level grounding requires the entities mentioned in a response to be present in the image. Reliable relation prediction goes further, requiring the model to combine evidence from the recognized entities and determine how they are related. For example, recognizing the man, cake, and candles does not by itself determine whether the man is blowing out the candles (Figure~\ref{fig:teaser}, left). An MLLM may nevertheless predict such an image-unsupported relation, producing a \textit{relation hallucination}~\cite{zheng2025reefknot,jiang2024hal,wu2025mitigating,qin2026mitigating,wu2026mmra}.

A common strategy for improving visual grounding is to strengthen or recalibrate visual signals during decoding~\cite{huang2024opera,yin2025clearsight,liu2024paying,yu2025visual,zhuang2025vasparse}. This is an intuitive remedy when a model overlooks an object. Relation prediction, however, requires evidence from multiple regions and may not improve simply because the image receives more attention. Under the same visual-attention amplification, performance improves steadily on the object-level POPE benchmark~\cite{Li-hallucination-2023} but changes only modestly and non-monotonically on the relation-focused Reefknot benchmark~\cite{zheng2025reefknot} (Section~\ref{sec:pre}). This contrast motivates examining how visual evidence is used across decoding, beyond overall visual attention strength.

We therefore trace visual attention in relation descriptions derived from Reefknot, comparing how visual focus evolves in correct and hallucinated predictions (Section~\ref{sec:pre}). Regions that become dominant early in hallucinated predictions often remain dominant at middle and later output positions, whereas correct predictions show larger shifts in the emphasized regions. This contrast appears in individual examples and in group-level comparisons of attention displacement across output positions. The persistent reuse of previously dominant visual evidence reflects a form of \textit{visual inertia} (Figure~\ref{fig:teaser}, middle). To test how stronger persistence affects relation prediction, we progressively carry more previous-step visual attention into the current step. Performance steadily deteriorates as carryover increases, indicating that stronger temporal persistence can impair relation prediction.

Guided by this temporal pattern, we introduce Inertia-aware Visual Excitation (\ours), which couples reinforcement of newly relevant evidence with conditional attenuation of persistently dominant contributions. Recent attention history distinguishes emergent tokens with rising attention from inertia tokens that remain strongly attended without a sufficient relative rise. For inertia tokens, attenuation reflects how frequently they have previously occupied this role. These roles are reassessed as generation proceeds, allowing a previously dominant token to be reinforced when its attention rises again relative to history. The method recalibrates visual values during decoding without additional training or parameter updates.

Across three MLLM backbones, \ours consistently improves relation prediction on Reefknot~\cite{zheng2025reefknot} and Hal-Eval~\cite{jiang2024hal}. Results on MMRA~\cite{wu2026mmra} extend this benefit to a multi-image relational setting. \ours also remains effective across decoding strategies while preserving performance on broader multimodal benchmarks.

\section{From Attention Strength to Temporal Persistence}
\label{sec:pre}

Many hallucination-mitigation methods strengthen visual attention to improve grounding~\cite{yin2025clearsight,liu2024paying,yu2025visual}. For relation prediction, however, the image must provide evidence not only for the entities mentioned in a response but also for how they are related. A model may attend strongly to the relevant entities while continuing to emphasize evidence that supports their recognition rather than the predicted relation.

\subsection{More Visual Attention Is Not Enough}

To examine whether stronger visual attention is sufficient to improve relation prediction, we vary the amplification factor on LLaVA-1.5~\cite{liu2024improved} following PAI~\cite{liu2024paying}. We apply the same settings to POPE~\cite{Li-hallucination-2023}, which evaluates object hallucination, and Reefknot~\cite{zheng2025reefknot}, which evaluates relation hallucination.

\begin{wrapfigure}{r}{0.54\textwidth}
    \centering
    \vspace{-8mm}
    \includegraphics[width=\linewidth]{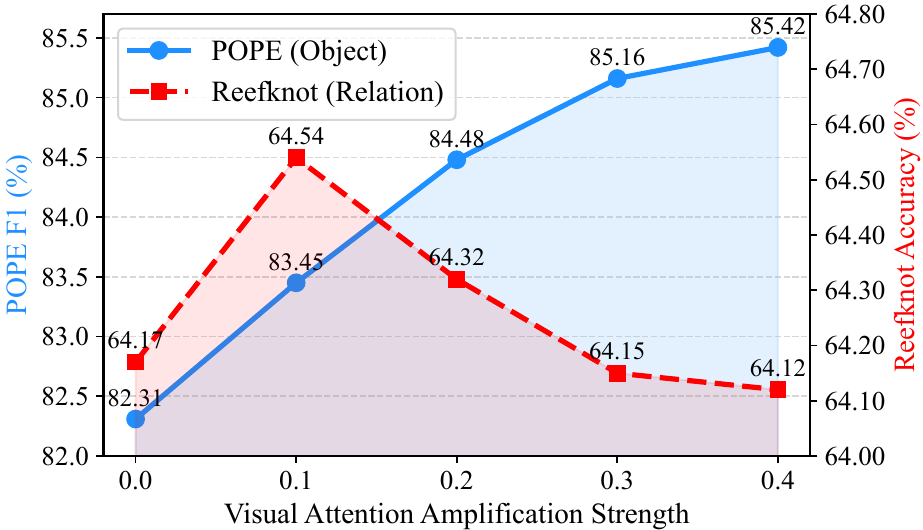}
    \vspace{-8mm}
    \caption{\textbf{Visual-attention amplification.} POPE improves steadily, while Reefknot gains only under mild amplification before returning to its baseline range.}
    \label{fig:amplify}
    \vspace{-12mm}
\end{wrapfigure}

POPE F1 improves steadily by 3.11 points, whereas Reefknot gains only under mild amplification before returning to its baseline range (Figure~\ref{fig:amplify}). Stronger visual attention thus benefits object-level evaluation without consistently improving relation prediction. This contrast motivates examining how visual attention evolves during generation, beyond its overall strength.

\subsection{Visual Evidence Remains Persistent Across Decoding}

\begin{figure}[t]
    \centering
    \includegraphics[width=\linewidth]{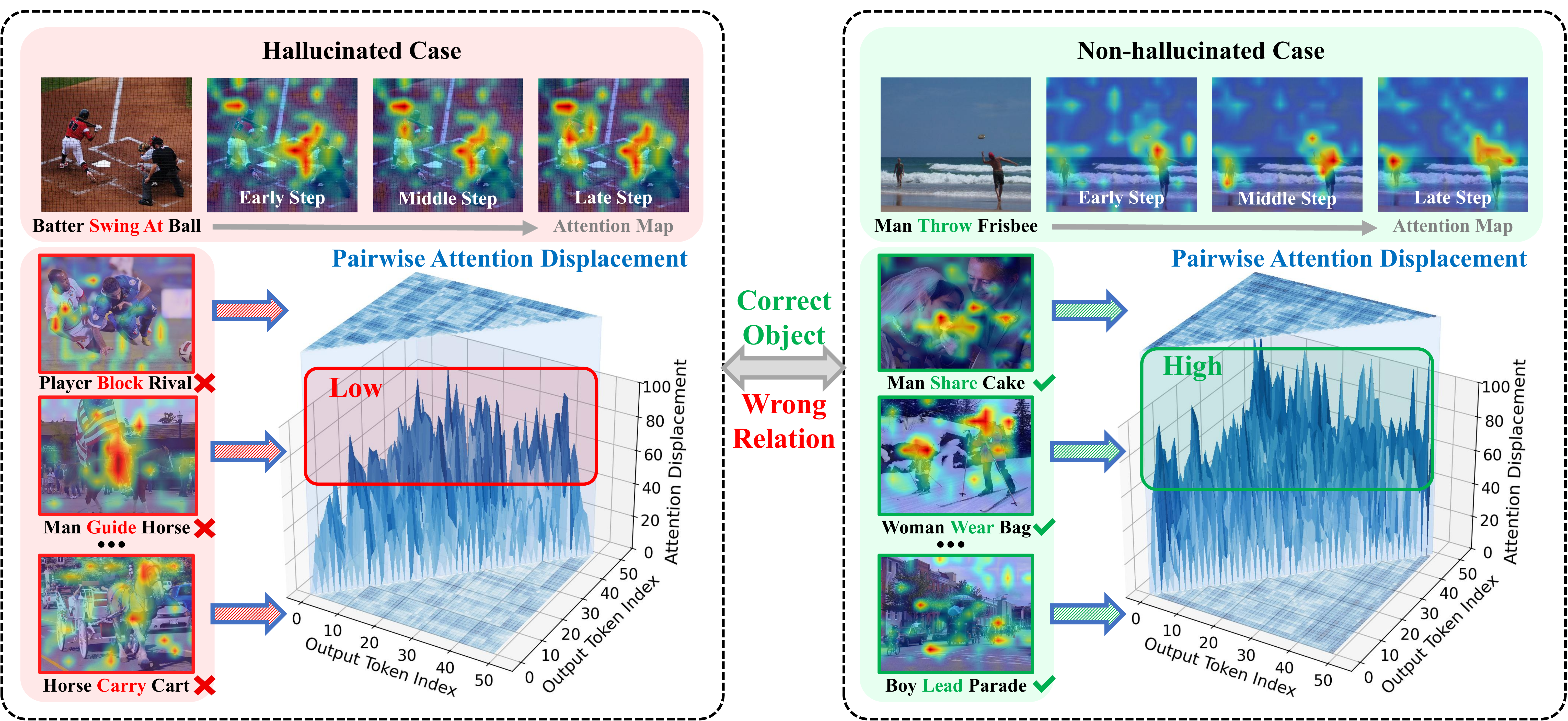}
    \vspace{-7mm}
    \caption{\textbf{Visual-attention persistence in Reefknot-derived relation descriptions.} Early, middle, and late attention maps illustrate hallucinated (left) and correct (right) descriptions. The surfaces show each group's mean pairwise index-based attention displacement. Lower values indicate more similar visual-attention distributions across output positions, with the hallucinated group showing more persistent visual focus.}
    \label{fig:attention_trajectories}
\end{figure}

We analyze visual-attention trajectories in relation descriptions derived from Reefknot, using the generation and annotation protocol in Appendix~\ref{sec:relation_description_protocol}. Across early, middle, and late attention maps (Figure~\ref{fig:attention_trajectories}), attention shifts between regions in the correct example but stays on similar regions in the hallucinated example. This continued reliance on previously dominant visual evidence illustrates visual inertia across decoding.

To characterize persistence beyond these snapshots, we analyze 1,000 descriptions under the same Vanilla configuration, with 500 correct, non-hallucinated descriptions and 500 hallucinated descriptions. For each response, we compare attention distributions across all pairs of analyzed output positions, capturing whether visual focus remains similar over both short and long intervals.

To compare visual-token allocation, we aggregate over $N_h$ heads and normalize the visual slice. With $\mathbf{A}^{(h)}_{t,\mathcal{V}}$ denoting attention from query $t$ to visual tokens $\mathcal{V}$ in head $h$ of the analyzed layer, we obtain
\begin{equation}
    \mathbf{p}_t =
    \frac{\sum_{h=1}^{N_h}\mathbf{A}^{(h)}_{t,\mathcal{V}}}
    {\left\|\sum_{h=1}^{N_h}\mathbf{A}^{(h)}_{t,\mathcal{V}}\right\|_1}.
    \label{eq:normalized_visual_attention}
\end{equation}
For prediction $n$, the resulting pairwise displacement matrix is
\begin{equation}
    \mathbf{M}^{(n)}_{i,j}=W_{\mathrm{idx}}\!\left(\mathbf{p}^{(n)}_i,\mathbf{p}^{(n)}_j\right),
    \qquad 1\leq i,j\leq 50,
    \label{eq:pairwise_displacement}
\end{equation}
where $W_{\mathrm{idx}}$ is the one-dimensional Wasserstein distance on row-major patch indices $a,b$ with cost $|a-b|$, measuring displacement in index units. Larger values indicate greater attention change, whereas extended low-value regions indicate similar distributions across output positions. Averaging per-prediction matrices elementwise within each group yields the surfaces in Figure~\ref{fig:attention_trajectories}.

Correct descriptions exhibit larger displacement across output-position pairs, whereas hallucinated descriptions retain more similar indexed attention distributions. In these descriptions, relation errors are associated with more persistent visual focus. A prediction-level analysis in Appendix~\ref{sec:consecutive_displacement} likewise finds higher relation accuracy among trajectories with larger consecutive attention displacement.

\subsection{Increasing Temporal Persistence Degrades Relation Prediction}

Hallucinated descriptions in the preceding analysis tend to retain similar visual focus across decoding positions. We examine how relation prediction changes when previous-step attention is explicitly carried forward. On LLaVA-1.5 and Reefknot, we progressively mix previous-step attention into the current distribution, with $\lambda$ controlling the previous step's contribution:
\begin{equation}
    \mathbf{p}^{\mathrm{carry}}_t(\lambda)
    =\operatorname{Norm}\!\left(\mathbf{p}_t+\lambda\mathbf{p}_{t-1}\right).
    \label{eq:attention_carryover}
\end{equation}

\begin{wraptable}{r}{0.45\textwidth}
    \centering
    \vspace{-7mm}
    \caption{\textbf{Progressive carryover on Reefknot Y/N.} The reference matches carryover entropy using current-step attention only. $D_{\mathrm{attn}}$ measures consecutive displacement.}
    \label{tab:progressive_carryover}
    \footnotesize
    \begin{tabular*}{\linewidth}{@{\extracolsep{\fill}}ccccc@{}}
        \toprule
        & \multicolumn{2}{c}{\textbf{Carryover}} & \multicolumn{2}{c}{\textbf{Matched reference}} \\
        \cmidrule(lr){2-3}\cmidrule(lr){4-5}
        $\lambda$ & $D_{\mathrm{attn}}$ & $R_{score}\uparrow$ & $D_{\mathrm{attn}}$ & $R_{score}\uparrow$ \\
        \midrule
        0.00 & 41.52 & 64.39 & 41.52 & 64.39 \\
        0.25 & 35.41 & 59.61 & 38.13 & 62.27 \\
        0.50 & 28.73 & 54.50 & 34.62 & 59.74 \\
        0.75 & 23.64 & 50.71 & 31.17 & 57.55 \\
        1.00 & 19.24 & 47.68 & 28.84 & 55.86 \\
        \bottomrule
    \end{tabular*}
    \vspace{-5mm}
\end{wraptable}

Carrying forward previous attention can also change the distribution's concentration. We therefore pair each setting with an entropy-matched reference built solely from current-step attention, comparing carryover and current-step reshaping at matched concentration (Appendix~\ref{sec:carryover_control}).

As carryover strengthens, $R_{score}$ steadily declines, with a larger drop than under the entropy-matched reference at every nonzero strength (Table~\ref{tab:progressive_carryover}). The performance gap widens as carryover increases, accompanied by a sharper reduction in consecutive attention displacement.

\section{Inertia-aware Visual Excitation}
\label{sec:approach}

\begin{figure}[t]
    \centering
    \includegraphics[width=\linewidth]{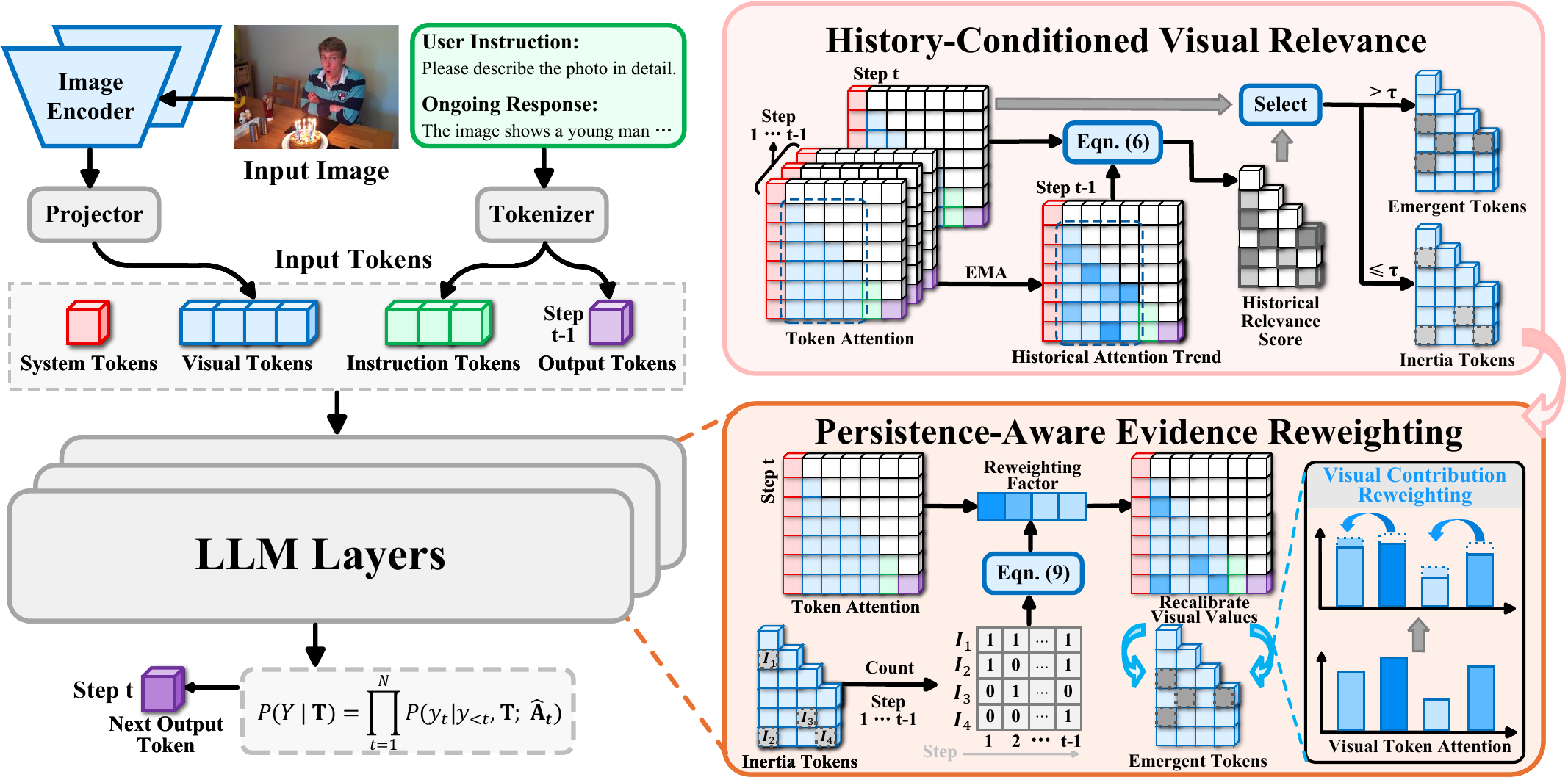}
    \vspace{-7mm}
    \caption{\textbf{Overview of \ours.} The method compares current visual attention with recent history to identify emergent and inertia tokens. Recalibration relative to their original visual values reinforces newly relevant evidence and mildly attenuates persistently dominant contributions during decoding.}
    \label{fig:method}
\end{figure}

The persistence observed in relation descriptions motivates calibrating visual contributions using attention history rather than current strength alone (Section~\ref{sec:pre}). Inertia-aware Visual Excitation (\ours) combines token-specific historical references with emergent-token reinforcement and conditional inertia-token attenuation. Figure~\ref{fig:method} outlines this training-free method.

\subsection{History-Conditioned Visual Relevance}

Consider two visual tokens that receive the same attention at the current decoding step. One may have remained at this level for several steps, while the other may have only recently risen from a weak historical level. A current-only signal treats them identically, although the latter reflects a change in visual focus. \ours preserves this distinction by evaluating each visual token relative to its own history.

To track recent changes in each token's attention, we maintain an exponential moving average as a short-term reference. At selection layer $l_s$, we read the attention vector $\mathbf{A}^{(h)}_t\in\mathbb{R}^{N}$ over all $N$ tokens before value recalibration. Averaging the $N_h$ heads gives $a_{t,j}$ for visual token $j\in\mathcal{V}$, with its historical reference updated as
\begin{align}
    a_{t,j}
    &=\frac{1}{N_h}\sum_{h=1}^{N_h}\mathbf{A}^{(h)}_{t,j},
    \label{eq:head_averaged_attention}\\
    \widetilde{a}_{t-1,j}
    &=\gamma\widetilde{a}_{t-2,j}+(1-\gamma)a_{t-1,j}.
    \label{eq:attention_ema}
\end{align}
The reference blends the previous observation and the older moving average with weights $1-\gamma$ and $\gamma\in(0,1)$, respectively. Here, $t$ indexes output-predicting queries, with $t=1$ at the final prompt position. The recurrence applies for $t>1$, starting from $\widetilde a_{0,j}$.

Raw attention changes can differ in scale across visual tokens. We therefore compare the current signal with its historical reference using
\begin{equation}
    S_{t,j}=
    \frac{a_{t,j}-\widetilde{a}_{t-1,j}}
    {\sqrt{\widetilde{a}_{t-1,j}(1-\widetilde{a}_{t-1,j})}+\epsilon},
    \qquad \epsilon=10^{-6}.
    \label{eq:history_relevance}
\end{equation}
This normalized difference preserves the direction of change and accounts for historical attention scale. The signal reflects changes in both overall visual attention and its allocation across visual tokens. A large positive $S_{t,j}$ indicates increased prominence relative to the token's recent history.

Emergent tokens show a sufficient rise relative to history, while inertia tokens remain prominent with limited relative growth. To identify the latter, we combine the relevance score with current attention strength. Using the visual-token mean $\mu_t^v=N_v^{-1}\sum_{j\in\mathcal{V}}a_{t,j}$, where $N_v=|\mathcal{V}|$, we define
\begin{equation}
\begin{aligned}
    \mathcal{E}_t
    &=\{j\in\mathcal{V}\mid S_{t,j}>\tau\}, \\
    \mathcal{I}_t
    &=\{j\in\mathcal{V}\mid a_{t,j}>\mu_t^v,\;S_{t,j}\leq\tau\},
\end{aligned}
    \label{eq:visual_token_sets}
\end{equation}
where $\tau$ is the relevance threshold. The above-mean condition focuses inertia selection on prominent contributions, distinguishing them from weakly attended tokens with similarly low scores.

\subsection{Persistence-Aware Evidence Reweighting}

The inertia set determines which visual contributions are eligible for attenuation. To distinguish repeated dominance from an occasional selection, \ours sets attenuation strength using prior inertia-selection frequency. With initial count $C_j^{\mathrm{pre}}$ and history length $n_{\mathrm{pre}}$, we accumulate the history before output token $t$ as
\begin{equation}
    C_{t,j}=C_j^{\mathrm{pre}}+\sum_{k=1}^{t-1}\mathbf{1}[j\in\mathcal{I}_k],
    \qquad n_t=n_{\mathrm{pre}}+t-1.
    \label{eq:persistence_count}
\end{equation}
The ratio $C_{t,j}/n_t$ measures historical inertia-selection frequency, with higher values producing stronger attenuation for currently selected inertia tokens. The current query is excluded, and tokens without prior inertia selections receive no attenuation. The modulation factor before capping is
\begin{equation}
    m_{t,j}=
    \begin{cases}
        1-\alpha\dfrac{C_{t,j}}{n_t},
        & j\in\mathcal{I}_t,\ n_t>0, \\
        1+S_{t,j},
        & j\in\mathcal{E}_t, \\
        1,
        & \text{otherwise},
    \end{cases}
    \label{eq:modulation_factor}
\end{equation}
where $\alpha\in(0,1]$ controls attenuation, with inertia-token factors bounded by $[1-\alpha,1]$. Emergent-token selection uses the original relevance score, while its multiplier is capped at $1.10$, yielding fixed-strength reinforcement at the reported thresholds.

The token sets are reassessed at each step, so renewed attention can move an inertia token into the emergent set for reinforcement. These query-specific factors scale each token's contribution through its visual value. Let $\mathbf{V}^{(l,h),\mathrm{base}}_j$ denote the fixed original value at layer $l$ and head $h$. Using the capped factor, we apply
\begin{equation}
    \mathbf{V}^{(l,h)}_j
    \leftarrow m_{t,j}\mathbf{V}^{(l,h),\mathrm{base}}_j.
    \label{eq:value_recalibration}
\end{equation}
Tokens outside both sets regain their original values. Recalibration spans all layers and heads once per query, before attention--value aggregation at $l_s$. The adjusted values affect $l_s$ and later layers immediately, and earlier layers on subsequent queries. Algorithm~\ref{alg:ive} details execution and history updates (Appendix~\ref{sec:implementation}).

\section{Experiments}
\label{exp}

We evaluate whether \ours reduces relation hallucination across three MLLMs while preserving performance on broader multimodal tasks.
We also examine component contributions, robustness across configurations and decoding strategies, and computational cost.

\begin{table}[t]
\centering
\caption{Results (\%) on Reefknot under Y/N and MCQ evaluation. Perc. and Cogn. denote the benchmark's perception- and cognition-hallucination rates.}
\label{tab:reefknot}
\resizebox{\linewidth}{!}{%
\begin{tabular}{clccccccccc}
\toprule
& &
\multicolumn{3}{c}{\textbf{LLaVA-1.5}} &
\multicolumn{3}{c}{\textbf{InstructBLIP}} &
\multicolumn{3}{c}{\textbf{Qwen2.5-VL-Instruct}} \\
\cmidrule(lr){3-5}
\cmidrule(lr){6-8}
\cmidrule(lr){9-11}
\textbf{Category} & \textbf{Method} & \textbf{Perc.$\downarrow$} & \textbf{Cogn.$\downarrow$} & $\mathbf{R_{score}\uparrow}$ & \textbf{Perc.$\downarrow$} & \textbf{Cogn.$\downarrow$} & $\mathbf{R_{score}\uparrow}$ & \textbf{Perc.$\downarrow$} & \textbf{Cogn.$\downarrow$} & $\mathbf{R_{score}\uparrow}$ \\
\midrule
& Vanilla &
37.67 & 33.99 & 64.39 &
45.31 & 42.47 & 56.28 &
37.22 & 20.58 & 72.07 \\
\addlinespace[1pt]
& \multicolumn{10}{c}{\cellcolor[HTML]{F0F0F0}\textbf{Object-specific}} \\
& +VCD~\cite{leng2024mitigating} &
37.33 & 33.67 & 64.71 &
45.22 & 42.35 & 56.38 &
37.13 & 20.47 & 72.17 \\
\addlinespace[1pt]
& \multicolumn{10}{c}{\cellcolor[HTML]{F0F0F0}\textbf{General}} \\
& +OPERA~\cite{huang2024opera} &
37.08 & 33.23 & 65.07 &
45.03 & 42.05 & 56.63 &
36.86 & 20.12 & 72.49 \\
& +PAI~\cite{liu2024paying} &
37.21 & 33.77 & 64.71 &
44.89 & 41.86 & 56.80 &
36.82 & 19.95 & 72.60 \\
& +HELPD~\cite{yuan2024helpd} &
37.05 & 33.31 & 65.04 &
44.58 & 41.94 & 56.89 &
36.79 & 20.04 & 72.57 \\
& +DeCo~\cite{wang2024mllmseedynamiccorrection} &
37.22 & 33.42 & 64.90 &
45.17 & 42.21 & 56.48 &
37.00 & 20.33 & 72.31 \\
& +VASparse~\cite{zhuang2025vasparse} &
37.13 & 33.14 & 65.10 &
44.67 & 42.47 & 56.56 &
36.91 & 20.86 & 72.05 \\
& +MemVR~\cite{zou2024look} &
36.99 & 32.81 & 65.34 &
44.34 & 42.09 & 56.92 &
36.75 &  20.16 & 72.52 \\
\addlinespace[1pt]
& \multicolumn{10}{c}{\cellcolor[HTML]{F0F0F0}\textbf{Relation-specific}} \\
& +CAP~\cite{wu2025mitigating} &
37.32 & 33.62 & 64.75 &
44.97 & 42.18 & 56.59 &
36.94 & 20.31 & 72.35 \\
& +DtC~\cite{zheng2025reefknot} &
36.94 & 32.87 & 65.33 &
44.63 & 41.82 & 56.94 &
36.71 & 20.08 & 72.58 \\
& +RVE~\cite{qin2026mitigating} &
37.06 & 33.04 & 65.19 &
44.79 & 42.01 & 56.76 &
36.97 & 20.15 & 72.42 \\
& \cellcolor[HTML]{E8F2FE}\textbf{+IVE (Ours)} &
\textbf{34.92} & \textbf{31.11} & \textbf{67.21} &
\textbf{42.94} & \textbf{39.31} & \textbf{59.09} &
\textbf{34.91} & \textbf{17.60} & \textbf{74.76} \\
\multirow{-16}{*}{Y/N} & \cellcolor[HTML]{E8F2FE}\textbf{$\Delta$ vs. Vanilla} &
\cellcolor[HTML]{E8F2FE}{\color[HTML]{CB0000}\textbf{-2.75}} & \cellcolor[HTML]{E8F2FE}{\color[HTML]{CB0000}\textbf{-2.88}} &
\cellcolor[HTML]{E8F2FE}{\color[HTML]{CB0000}\textbf{+2.82}} & \cellcolor[HTML]{E8F2FE}{\color[HTML]{CB0000}\textbf{-2.37}} &
\cellcolor[HTML]{E8F2FE}{\color[HTML]{CB0000}\textbf{-3.16}} & \cellcolor[HTML]{E8F2FE}{\color[HTML]{CB0000}\textbf{+2.81}} &
\cellcolor[HTML]{E8F2FE}{\color[HTML]{CB0000}\textbf{-2.31}} & \cellcolor[HTML]{E8F2FE}{\color[HTML]{CB0000}\textbf{-2.98}} &
\cellcolor[HTML]{E8F2FE}{\color[HTML]{CB0000}\textbf{+2.69}} \\
\midrule
& Vanilla &
68.05 & 51.04 & 43.70 &
73.91 & 53.85 & 39.94 &
67.16 & 49.69 & 44.91 \\
\addlinespace[1pt]
& \multicolumn{10}{c}{\cellcolor[HTML]{F0F0F0}\textbf{Object-specific}} \\
& +VCD~\cite{leng2024mitigating} &
67.86 & 50.90 & 43.85 &
73.67 & 53.75 & 40.09 &
67.02 & 49.58 & 45.02 \\
\addlinespace[1pt]
& \multicolumn{10}{c}{\cellcolor[HTML]{F0F0F0}\textbf{General}} \\
& +OPERA~\cite{huang2024opera} &
67.12 & 50.21 & 44.56 &
72.98 & 53.23 & 40.66 &
66.51 & 49.13 & 45.49 \\
& +PAI~\cite{liu2024paying} &
66.93 & 51.10 & 44.00 &
74.05 & 53.83 & 39.91 &
67.58 & 49.25 & 45.08 \\
& +HELPD~\cite{yuan2024helpd} &
66.84 & 50.58 & 44.39 &
73.18 & 53.27 & 40.57 &
66.44 & 49.04 & 45.58 \\
& +DeCo~\cite{wang2024mllmseedynamiccorrection} &
67.54 & 50.73 & 44.07 &
73.30 & 53.60 & 40.31 &
66.84 & 49.38 & 45.22 \\
& +VASparse~\cite{zhuang2025vasparse} &
67.21 & 50.88 & 44.07 &
73.91 & 53.48 & 40.20 &
66.51 & 49.21 & 45.44 \\
& +MemVR~\cite{zou2024look} &
66.98 & 50.79 & 44.20 &
73.40 & 53.19 & 40.56 &
66.37 & 49.15 & 45.52 \\
\addlinespace[1pt]
& \multicolumn{10}{c}{\cellcolor[HTML]{F0F0F0}\textbf{Relation-specific}} \\
& +CAP~\cite{wu2025mitigating} &
67.40 & 50.81 & 44.06 &
73.49 & 53.21 & 40.52 &
66.79 & 49.33 & 45.27 \\
& +DtC~\cite{zheng2025reefknot} &
66.70 & 50.50 & 44.49 &
72.74 & 52.88 & 40.98 &
66.19 & 49.08 & 45.63 \\
& +RVE~\cite{qin2026mitigating} &
66.93 & 50.63 & 44.33 &
73.07 & 52.75 & 40.96 &
66.56 & 49.02 & 45.55 \\
& \cellcolor[HTML]{E8F2FE}\textbf{+IVE (Ours)} &
\textbf{65.02} & \textbf{49.08} & \textbf{45.99} &
\textbf{70.89} & \textbf{50.87} & \textbf{42.94} &
\textbf{64.75} & \textbf{47.01} & \textbf{47.50} \\
\multirow{-16}{*}{MCQ} & \cellcolor[HTML]{E8F2FE}\textbf{$\Delta$ vs. Vanilla} &
\cellcolor[HTML]{E8F2FE}{\color[HTML]{CB0000}\textbf{-3.03}} & \cellcolor[HTML]{E8F2FE}{\color[HTML]{CB0000}\textbf{-1.96}} &
\cellcolor[HTML]{E8F2FE}{\color[HTML]{CB0000}\textbf{+2.29}} & \cellcolor[HTML]{E8F2FE}{\color[HTML]{CB0000}\textbf{-3.02}} &
\cellcolor[HTML]{E8F2FE}{\color[HTML]{CB0000}\textbf{-2.98}} & \cellcolor[HTML]{E8F2FE}{\color[HTML]{CB0000}\textbf{+3.00}} &
\cellcolor[HTML]{E8F2FE}{\color[HTML]{CB0000}\textbf{-2.41}} & \cellcolor[HTML]{E8F2FE}{\color[HTML]{CB0000}\textbf{-2.68}} &
\cellcolor[HTML]{E8F2FE}{\color[HTML]{CB0000}\textbf{+2.59}} \\
\bottomrule
\end{tabular}
}
\vspace{-5mm}
\end{table}

\subsection{Experimental Settings}

\textbf{Benchmarks: }
Reefknot~\cite{zheng2025reefknot} and Hal-Eval~\cite{jiang2024hal} provide our primary evaluation of relation hallucination.
Reefknot covers Y/N and multiple-choice relation prediction (scoring in Appendix~\ref{sec:implementation}), while Hal-Eval distinguishes relation, object, attribute, and event hallucinations.
MMRA~\cite{wu2026mmra} evaluates multi-granularity, multi-image relational association.
POPE~\cite{Li-hallucination-2023} provides an object-hallucination comparison.
We use MMBench~\cite{liu2024mmbench} and MME~\cite{fu2024mmecomprehensiveevaluationbenchmark} to assess whether \ours preserves performance on broader multimodal tasks.

\noindent\textbf{Models: }
We evaluate \ours on the 7B variants of LLaVA-1.5~\cite{liu2024improved}, InstructBLIP~\cite{dai2023instructblip}, and Qwen2.5-VL-Instruct~\cite{bai2025qwen2}.

\noindent\textbf{Baselines: }
Besides Vanilla, we compare with object-specific VCD~\cite{leng2024mitigating}.
General methods include OPERA~\cite{huang2024opera}, PAI~\cite{liu2024paying}, HELPD~\cite{yuan2024helpd}, DeCo~\cite{wang2024mllmseedynamiccorrection}, VASparse~\cite{zhuang2025vasparse}, and MemVR~\cite{zou2024look}.
Relation-specific baselines include CAP~\cite{wu2025mitigating}, DtC~\cite{zheng2025reefknot}, and RVE~\cite{qin2026mitigating}.
CAP, DtC, and RVE are evaluated on both Reefknot settings and Hal-Eval Relation, with RVE reproduced from its paper description.

\noindent\textbf{Implementation Details: }
We use selection layers 20, 20, and 18 for the three backbones and recalibrate visual values across all layers and heads within each decoding forward pass (Appendix~\ref{sec:implementation}).
Selection-layer ablations appear in Appendix Table~\ref{tab:layer_ablation}.
Main results use model-specific configurations selected on held-out Reefknot validation data and fixed across benchmarks before final evaluation.
Experiments run on NVIDIA H800 GPUs.

\begin{table}[t]
\centering
\caption{Relation-type results (\%) on Hal-Eval across three MLLMs.}
\label{tab:haleval}
\resizebox{\linewidth}{!}{%
\begin{tabular}{lcccccccccccc}
\toprule
\multirow{2}{*}{\textbf{Method}} &
\multicolumn{4}{c}{\textbf{LLaVA-1.5}} &
\multicolumn{4}{c}{\textbf{InstructBLIP}} &
\multicolumn{4}{c}{\textbf{Qwen2.5-VL-Instruct}} \\ 
\cmidrule(l){2-5} 
\cmidrule(l){6-9} 
\cmidrule(l){10-13} 
& \textbf{Acc}$\uparrow$ & $\mathbf{\Delta}$ & \textbf{F1}$\uparrow$ & $\mathbf{\Delta}$ & \textbf{Acc}$\uparrow$ & $\mathbf{\Delta}$ & \textbf{F1}$\uparrow$ & $\mathbf{\Delta}$ & \textbf{Acc}$\uparrow$ & $\mathbf{\Delta}$ & \textbf{F1}$\uparrow$ & $\mathbf{\Delta}$ \\ 
\midrule

Vanilla &
52.65 & +0.00 & 35.82 & +0.00 &
49.80 & +0.00 & 66.35 & +0.00 &
62.43 & +0.00 & 57.28 & +0.00 \\

\addlinespace[1pt]
\multicolumn{13}{c}{\cellcolor[HTML]{F0F0F0}\textbf{Object-specific}} \\
+VCD~\cite{leng2024mitigating} &
53.91 & +1.26 & 37.08 & +1.26 &
50.88 & +1.08 & 67.06 & +0.71 &
63.37 & +0.94 & 58.26 & +0.98 \\

\addlinespace[1pt]
\multicolumn{13}{c}{\cellcolor[HTML]{F0F0F0}\textbf{General}} \\
+OPERA~\cite{huang2024opera} &
55.47 & +2.82 & 38.53 & +2.71 &
51.73 & +1.93 & 67.41 & +1.06 &
64.95 & +2.52 & 59.43 & +2.15 \\

+PAI~\cite{liu2024paying} &
55.12 & +2.47 & 38.02 & +2.20 &
51.38 & +1.58 & 67.29 & +0.94 &
64.62 & +2.19 & 59.08 & +1.80 \\

+HELPD~\cite{yuan2024helpd} &
55.41 & +2.76 & 38.46 & +2.64 &
51.94 & +2.14 & 67.54 & +1.19 &
65.18 & +2.75 & 59.97 & +2.69 \\

+DeCo~\cite{wang2024mllmseedynamiccorrection} &
54.76 & +2.11 & 37.91 & +2.09 &
51.21 & +1.41 & 67.21 & +0.86 &
64.18 & +1.75 & 58.97 & +1.69 \\

+VASparse~\cite{zhuang2025vasparse} &
55.34 & +2.69 & 38.61 & +2.79 &
52.27 & +2.47 & 67.69 & +1.34 &
65.41 & +2.98 & 60.42 & +3.14 \\

+MemVR~\cite{zou2024look} &
55.83 & +3.18 & 39.04 & +3.22 &
52.61 & +2.81 & 67.78 & +1.43 &
65.62 & +3.19 & 60.83 & +3.55 \\

\addlinespace[1pt]
\multicolumn{13}{c}{\cellcolor[HTML]{F0F0F0}\textbf{Relation-specific}} \\
+CAP~\cite{wu2025mitigating} &
54.71 & +2.06 & 37.96 & +2.14 &
51.68 & +1.88 & 67.42 & +1.07 &
64.82 & +2.39 & 59.74 & +2.46 \\
+DtC~\cite{zheng2025reefknot} &
55.61 & +2.96 & 38.83 & +3.01 &
52.74 & +2.94 & 67.77 & +1.42 &
65.49 & +3.06 & 60.91 & +3.63 \\
+RVE~\cite{qin2026mitigating} &
55.38 & +2.73 & 39.17 & +3.35 &
52.49 & +2.69 & 67.79 & +1.44 &
65.73 & +3.30 & 60.68 & +3.40 \\
\cellcolor[HTML]{E8F2FE}{\textbf{+IVE (Ours)}} &
\cellcolor[HTML]{E8F2FE}{\color[HTML]{CB0000}\textbf{57.71}} &
\cellcolor[HTML]{E8F2FE}{\color[HTML]{CB0000}\textbf{+5.06}} &
\cellcolor[HTML]{E8F2FE}{\color[HTML]{CB0000}\textbf{41.29}} &
\cellcolor[HTML]{E8F2FE}{\color[HTML]{CB0000}\textbf{+5.47}} &
\cellcolor[HTML]{E8F2FE}{\color[HTML]{CB0000}\textbf{54.93}} &
\cellcolor[HTML]{E8F2FE}{\color[HTML]{CB0000}\textbf{+5.13}} &
\cellcolor[HTML]{E8F2FE}{\color[HTML]{CB0000}\textbf{68.93}} &
\cellcolor[HTML]{E8F2FE}{\color[HTML]{CB0000}\textbf{+2.58}} &
\cellcolor[HTML]{E8F2FE}{\color[HTML]{CB0000}\textbf{67.79}} &
\cellcolor[HTML]{E8F2FE}{\color[HTML]{CB0000}\textbf{+5.36}} &
\cellcolor[HTML]{E8F2FE}{\color[HTML]{CB0000}\textbf{62.84}} &
\cellcolor[HTML]{E8F2FE}{\color[HTML]{CB0000}\textbf{+5.56}} \\

\bottomrule
\end{tabular}
}
\vspace{-7mm}
\end{table}

\begin{figure*}[t]
    \centering
    \begin{minipage}[t]{0.44\textwidth}
        \vspace{0pt}
        \centering
        \captionof{table}{Accuracy (\%) on MMRA with Qwen2.5-VL-Instruct.}
        \label{tab:mmra}
        \vspace{1mm}
        \footnotesize
        \setlength{\tabcolsep}{3pt}
        \begin{tabularx}{\linewidth}{>{\raggedright\arraybackslash}Xc}
            \toprule
            \textbf{Method} & \textbf{Accuracy}$\uparrow$ \\
            \midrule
            Vanilla & 50.45 \\
            +VCD~\cite{leng2024mitigating} & 51.12 \\
            +OPERA~\cite{huang2024opera} & 51.38 \\
            +PAI~\cite{liu2024paying} & 51.27 \\
            +MemVR~\cite{zou2024look} & 52.16 \\
            \rowcolor[HTML]{E8F2FE}
            \textbf{+IVE (Ours)} & \textbf{55.37} \\
            \rowcolor[HTML]{E8F2FE}
            \textbf{$\Delta$ vs. Vanilla} & {\color[HTML]{CB0000}\textbf{+4.92}} \\
            \bottomrule
        \end{tabularx}
    \end{minipage}
    \hfill
    \begin{minipage}[t]{0.54\textwidth}
        \vspace{0pt}
        \centering
        \caption{Results comparison (\%) on MMBench.}
        \includegraphics[width=0.92\linewidth]{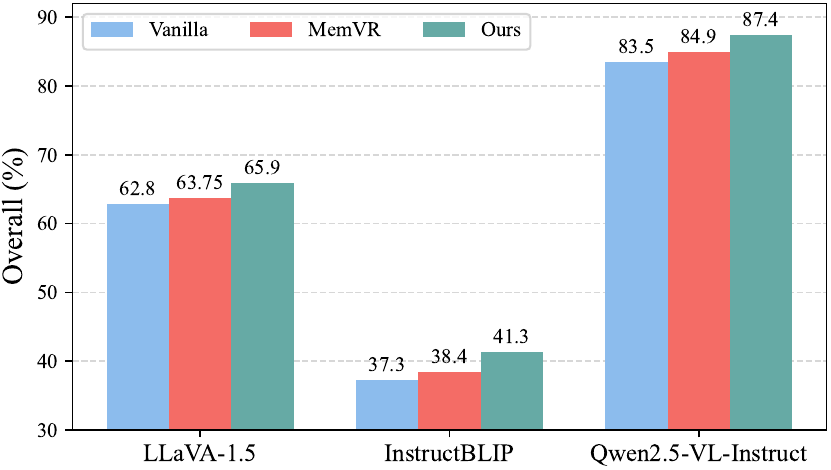}
        \label{fig:mmbench}
    \end{minipage}
    \vspace{-7mm}
\end{figure*}

\subsection{Main Results}

\textbf{Relation Hallucination: }
On Reefknot~\cite{zheng2025reefknot}, \ours attains the highest $R_{score}$ and lowest $Cogn.$ hallucination rates across all three backbones in both Y/N and MCQ settings (Table~\ref{tab:reefknot}).
These gains are accompanied by lower $Perc.$ hallucination rates.
The advantage over relation-specific baselines extends across these formats, covering both verification of a proposed relation and selection among alternatives.
\ours also outperforms ADAPT's training-free variant and Attention Never Lie across all three backbones in both Reefknot settings (Appendix Table~\ref{tab:temporal_method_results}).
On Hal-Eval~\cite{jiang2024hal}, \ours achieves higher Relation accuracy and F1 than all compared methods across the three backbones (Table~\ref{tab:haleval}).
For LLaVA-1.5, the 5.06-point accuracy gain over Vanilla exceeds the gains on Object, Attribute, and Event (Appendix Table~\ref{tab:haleval_types}).
The larger benefit on Relation is consistent with the method's focus on relational grounding.

\noindent\textbf{Relational Generalization on MMRA: }
To examine whether the gains extend to relations involving multiple images, we evaluate \ours on MMRA~\cite{wu2026mmra} with Qwen2.5-VL-Instruct.
\ours improves accuracy by 4.92 points over Vanilla and by 3.21 points over MemVR, the strongest baseline in this comparison (Table~\ref{tab:mmra}).
This result supports the use of history-conditioned visual reweighting when relation prediction draws on evidence from multiple images.

\noindent\textbf{Broader Multimodal Performance:}
\ours improves MMBench~\cite{liu2024mmbench} scores over Vanilla and MemVR across the three backbones (Figure~\ref{fig:mmbench}).
Scores on the selected MME categories also improve over Vanilla across all three models (Appendix Table~\ref{tab:mme}).
These results indicate preserved performance on the broader multimodal tasks evaluated here.
POPE and CHAIR further cover object-existence questions and generated captions, with CHAIR stratified by scene density and response length (Appendix Tables~\ref{tab:pope} and~\ref{tab:chair_stratified}).

\begin{table}[t]
\centering
\caption{Ablation, configuration robustness, and inference efficiency of \ours.}
\label{tab:ablation_efficiency}
\vspace{-3mm}
\begin{minipage}[t]{0.51\linewidth}
    \vspace{0pt}
    \centering
    {\small\textbf{(a) Component contributions}}\\[1mm]
    \vspace{-1mm}
    \scriptsize
    \setlength{\tabcolsep}{2.2pt}
    \resizebox{.9\linewidth}{!}{%
    \begin{tabular}{cccccc}
        \toprule
        \multirow{2}{*}{\textbf{Reinforce}} &
        \multirow{2}{*}{\textbf{Attenuate}} &
        \multicolumn{2}{c}{\textbf{Reefknot}} &
        \multicolumn{2}{c}{\textbf{POPE}} \\
        \cmidrule(lr){3-4}\cmidrule(lr){5-6}
        & & \textbf{Perc.}$\downarrow$ & \textbf{Cogn.}$\downarrow$ & \textbf{Acc.}$\uparrow$ & \textbf{F1}$\uparrow$ \\
        \midrule
        & & 37.67 & 33.99 & 83.29 & 81.33 \\
        $\checkmark$ & & 36.94 & 33.01 & 89.71 & 89.96 \\
        & $\checkmark$ & 37.12 & 33.24 & 88.96 & 89.21 \\
        \rowcolor[HTML]{E8F2FE}
        $\checkmark$ & $\checkmark$ & \textbf{34.92} & \textbf{31.11} & \textbf{90.53} & \textbf{90.94} \\
        \bottomrule
    \end{tabular}
    }
\end{minipage}
\hfill
\begin{minipage}[t]{0.45\linewidth}
    \vspace{0pt}
    \centering
    {\small\textbf{(b) Unified configuration}}\\[1mm]
    \vspace{-1mm}
    \small
    \setlength{\tabcolsep}{7pt}
    \resizebox{\linewidth}{!}{%
    \begin{tabular}{@{}lccc@{}}
        \toprule
        & \multicolumn{3}{c}{\textbf{Reefknot Y/N} ($\mathbf{R_{score}\uparrow}$)} \\
        \cmidrule(lr){2-4}
        \textbf{Backbone} & \textbf{Vanilla} & \textbf{Unified} & \textbf{Tuned} \\
        \midrule
        LLaVA-1.5 & 64.39 & 66.13 & \textbf{67.21} \\
        InstructBLIP & 56.28 & 57.93 & \textbf{59.09} \\
        Qwen2.5-VL & 72.07 & 73.38 & \textbf{74.76} \\
        \bottomrule
    \end{tabular}
    }
\end{minipage}

\vspace{1.5mm}
\begin{minipage}[t]{0.49\linewidth}
    \vspace{-3pt}
    \centering
    {\small\textbf{(c) Hyperparameter sensitivity}}\\[1mm]
    \vspace{-1mm}
    \scriptsize
    \setlength{\tabcolsep}{2.2pt}
    \resizebox{.95\linewidth}{!}{%
    \begin{tabular}{@{}cccccc@{}}
        \toprule
        \multirow{2}{*}{\textbf{Parameter}} & \multirow{2}{*}{\textbf{Value}} &
        \multicolumn{2}{c}{\textbf{Reefknot}} & \multicolumn{2}{c}{\textbf{POPE}} \\
        \cmidrule(lr){3-4}\cmidrule(lr){5-6}
        & & \textbf{Perc.}$\downarrow$ & \textbf{Cogn.}$\downarrow$ & \textbf{Acc.}$\uparrow$ & \textbf{F1}$\uparrow$ \\
        \midrule
        \multirow{3}{*}{$\alpha$}
        & 0.05 & 36.84 & 32.72 & 89.20 & 89.27 \\
        & \cellcolor[HTML]{E8F2FE}\textbf{0.10} & \cellcolor[HTML]{E8F2FE}\textbf{34.92} & \cellcolor[HTML]{E8F2FE}\textbf{31.11} & \cellcolor[HTML]{E8F2FE}\textbf{90.53} & \cellcolor[HTML]{E8F2FE}\textbf{90.94} \\
        & 0.15 & 36.60 & 32.43 & 89.57 & 89.62 \\
        \midrule
        \multirow{3}{*}{$\gamma$}
        & 0.08 & 36.56 & 32.48 & 89.77 & 89.82 \\
        & \cellcolor[HTML]{E8F2FE}\textbf{0.09} & \cellcolor[HTML]{E8F2FE}\textbf{34.92} & \cellcolor[HTML]{E8F2FE}\textbf{31.11} & \cellcolor[HTML]{E8F2FE}\textbf{90.53} & \cellcolor[HTML]{E8F2FE}\textbf{90.94} \\
        & 0.10 & 35.88 & 32.07 & 90.53 & 90.94 \\
        \midrule
        \multirow{3}{*}{$\tau$}
        & 2.95 & 36.72 & 32.44 & 89.63 & 89.68 \\
        & \cellcolor[HTML]{E8F2FE}\textbf{3.00} & \cellcolor[HTML]{E8F2FE}\textbf{34.92} & \cellcolor[HTML]{E8F2FE}\textbf{31.11} & \cellcolor[HTML]{E8F2FE}\textbf{90.53} & \cellcolor[HTML]{E8F2FE}\textbf{90.94} \\
        & 3.05 & 35.93 & 32.10 & 90.27 & 90.32 \\
        \bottomrule
    \end{tabular}
    }
\end{minipage}
\hfill
\begin{minipage}[t]{0.49\linewidth}
    \vspace{-3pt}
    \centering
    {\small\textbf{(d) Inference efficiency}}\\[1mm]
    \vspace{-1mm}
    \scriptsize
    \setlength{\tabcolsep}{2.8pt}
    \resizebox{.97\linewidth}{!}{%
    \begin{tabular}{@{}llccc@{}}
        \toprule
        \textbf{Backbone} & \textbf{Method} &  $\mathbf{R_{score}\uparrow}$ & \textbf{Time}$\downarrow$ & \textbf{Mem.}$\downarrow$ \\
        \midrule
        \multirow{4}{*}{LLaVA-1.5}
        & Vanilla & 64.39 & 0:14:44 & 15.15G \\
        & +VCD & 64.71 & 0:38:23 & 16.06G \\
        & +MemVR & 65.34 & 0:16:07 & 15.54G \\
        & \cellcolor[HTML]{E8F2FE}\textbf{+\ours} & \cellcolor[HTML]{E8F2FE}\textbf{67.21} & \cellcolor[HTML]{E8F2FE}\textbf{0:15:47} & \cellcolor[HTML]{E8F2FE}\textbf{15.38G} \\
        \midrule
        \multirow{4}{*}{InstructBLIP}
        & Vanilla & 56.28 & 0:47:02 & 31.13G \\
        & +VCD & 56.38 & 1:50:08 & 32.17G \\
        & +MemVR & 56.92 & 0:53:25 & 31.68G \\
        & \cellcolor[HTML]{E8F2FE}\textbf{+\ours} & \cellcolor[HTML]{E8F2FE}\textbf{59.09} & \cellcolor[HTML]{E8F2FE}\textbf{0:52:52} & \cellcolor[HTML]{E8F2FE}\textbf{31.45G} \\
        \midrule
        \multirow{4}{*}{Qwen2.5-VL}
        & Vanilla & 72.07 & 0:31:46 & 33.41G \\
        & +VCD & 72.17 & 1:32:21 & 34.65G \\
        & +MemVR & 72.52 & 0:34:02 & 33.72G \\
        & \cellcolor[HTML]{E8F2FE}\textbf{+\ours} & \cellcolor[HTML]{E8F2FE}\textbf{74.76} & \cellcolor[HTML]{E8F2FE}\textbf{0:33:45} & \cellcolor[HTML]{E8F2FE}\textbf{33.61G} \\
        \bottomrule
    \end{tabular}
    }
\end{minipage}
\vspace{-5mm}
\end{table}

\subsection{Ablation Studies and Efficiency}

\noindent\textbf{Component Contributions: }
We separately disable emergent-token reinforcement (\textit{Reinforce}) and inertia-token attenuation (\textit{Attenuate}) on LLaVA-1.5, retaining token-selection and history-update rules (Appendix~\ref{sec:component_ablation}).
Each branch improves over Vanilla on Reefknot~\cite{zheng2025reefknot} and POPE~\cite{Li-hallucination-2023}, while their combination performs best across the reported metrics (Table~\ref{tab:ablation_efficiency}(a)).
To examine the role of token selection within this paired design, we compare relevance signals under matched token budgets and fixed scaling factors. This matches token counts and scaling factors while varying which visual tokens receive the adjustments. Under these controls, the IVE score achieves the highest $R_{score}$ among tested signals (Appendix Table~\ref{tab:relevance_signal_comparison}).
Matched random-token and random-value controls yield smaller gains than structured selection and reweighting (Appendix Table~\ref{tab:random_controls}).

\noindent\textbf{Unified Configuration: }
To assess performance with a single configuration, we apply shared coefficients and a common layer-selection rule across the three backbones (Table~\ref{tab:ablation_efficiency}(b)).
We use $\alpha=0.10$, $\gamma=0.10$, $\tau=3.05$, and $l_{\mathrm{sel}}=\operatorname{round}(0.625L)$, where $L$ denotes the number of transformer layers in each backbone.
This unified configuration improves over Vanilla on all three backbones, while model-specific tuning yields further gains for each backbone.

\noindent\textbf{Hyperparameter Sensitivity: }
We vary $\alpha$, $\gamma$, and $\tau$ individually around the validation-selected LLaVA-1.5 setting while keeping the other coefficients fixed (Table~\ref{tab:ablation_efficiency}(c)).
This setting yields the lowest $Cogn.$ hallucination rate, but every tested setting improves over Vanilla on both Reefknot and POPE.
Gains persist throughout the tested neighborhood, with their magnitude varying across configurations.
Full three-backbone sensitivity results are reported in Appendix Table~\ref{tab:coefficient}.

\noindent\textbf{Inference Efficiency: }
On Reefknot, \ours improves $R_{score}$ over Vanilla with roughly 6--12\% longer runtime and less than 2\% additional GPU memory (Table~\ref{tab:ablation_efficiency}(d)).
It also achieves higher $R_{score}$ than VCD and MemVR across the three backbones, while requiring less runtime and memory than either.

\subsection{Analysis}

\noindent\textbf{Robustness Across Decoding Strategies: }
Because decoding strategies alter output sequences and visual-attention histories, we evaluate \ours under greedy decoding, beam search, and nucleus sampling on Reefknot with matched settings across methods.
Across all three backbones and strategies, \ours outperforms Vanilla and MemVR, with larger $R_{score}$ gains over Vanilla under greedy decoding than under beam search or sampling (Table~\ref{tab:decoding_robustness}).
Gains persist even when beam search improves Vanilla.

\begin{figure}[t]
    \centering
    \vspace{-3.5mm}
    \includegraphics[width=\linewidth]{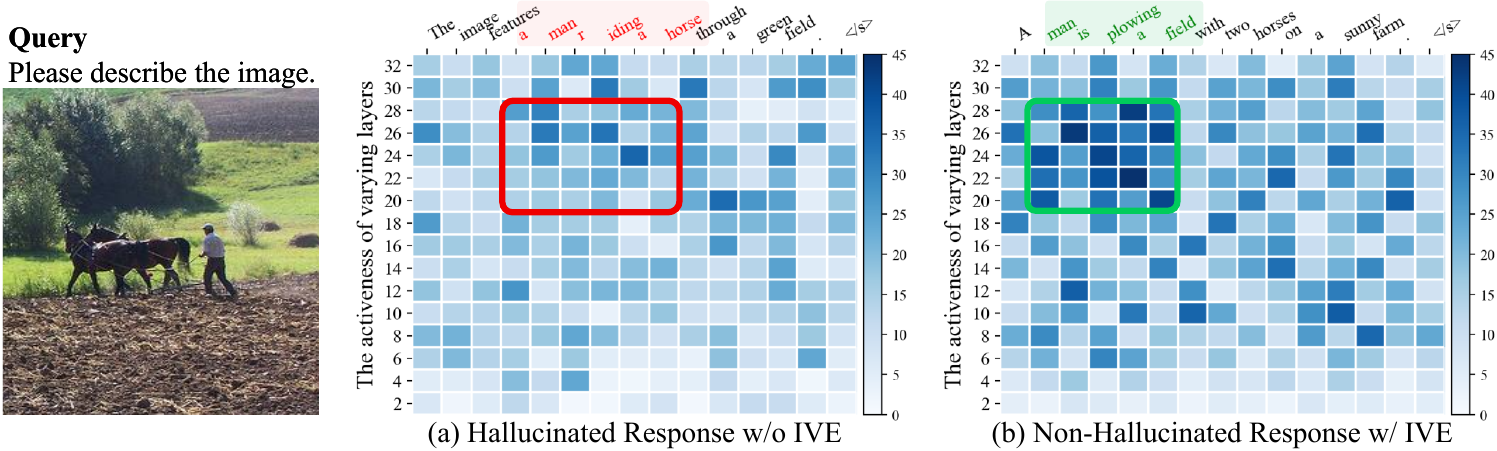}
    \vspace{-8.5mm}
    \caption{Layer-wise consecutive attention displacement for Vanilla and \ours on a Reefknot example. Rows denote LLaVA-1.5 layers and columns denote decoded tokens.}
    \vspace{-5mm}
    \label{fig:activeness_layers}
\end{figure}

\noindent\textbf{Layer-wise Attention Displacement: }
Figure~\ref{fig:activeness_layers} maps consecutive visual-attention displacement across layers and output positions for a representative LLaVA-1.5 Reefknot example.
This token-wise view localizes attention shifts across output positions.
Vanilla exhibits limited displacement around the unsupported description \textit{riding}, whereas \ours exhibits larger displacement around the supported description \textit{plowing a field}.
The contrast at relation-bearing positions reflects the persistent and shifting visual focus observed in Section~\ref{sec:pre}.
For LLaVA-1.5 and InstructBLIP, the method ordering is consistent under Wasserstein and Jensen--Shannon distances (Appendix Table~\ref{tab:distance_robustness}).

\begin{table}[t]
\centering
\caption{Y/N $R_{score}$ on Reefknot under different decoding strategies.}
\label{tab:decoding_robustness}
\resizebox{\linewidth}{!}{%
\begin{tabular}{lccccccccc}
\toprule
\multirow{2}{*}{\textbf{Decoding}} &
\multicolumn{3}{c}{\textbf{LLaVA-1.5}} &
\multicolumn{3}{c}{\textbf{InstructBLIP}} &
\multicolumn{3}{c}{\textbf{Qwen2.5-VL-Instruct}} \\
\cmidrule(lr){2-4}\cmidrule(lr){5-7}\cmidrule(lr){8-10}
& \textbf{Vanilla} & \textbf{MemVR} & \textbf{IVE}
& \textbf{Vanilla} & \textbf{MemVR} & \textbf{IVE}
& \textbf{Vanilla} & \textbf{MemVR} & \textbf{IVE} \\
\midrule
Greedy
& 64.39 & 65.34 & \cellcolor[HTML]{E8F2FE}\textbf{67.21}
& 56.28 & 56.92 & \cellcolor[HTML]{E8F2FE}\textbf{59.09}
& 72.07 & 72.52 & \cellcolor[HTML]{E8F2FE}\textbf{74.76} \\
Beam ($b=4$)
& 65.26 & 66.10 & \cellcolor[HTML]{E8F2FE}\textbf{67.10}
& 56.81 & 57.51 & \cellcolor[HTML]{E8F2FE}\textbf{58.57}
& 72.79 & 73.18 & \cellcolor[HTML]{E8F2FE}\textbf{74.20} \\
Sampling ($p=0.9$)
& $63.94{\pm}0.42$ & $64.85{\pm}0.45$ & \cellcolor[HTML]{E8F2FE}$\mathbf{65.77{\pm}0.39}$
& $55.96{\pm}0.51$ & $56.66{\pm}0.48$ & \cellcolor[HTML]{E8F2FE}$\mathbf{57.76{\pm}0.47}$
& $71.82{\pm}0.38$ & $72.24{\pm}0.41$ & \cellcolor[HTML]{E8F2FE}$\mathbf{73.32{\pm}0.35}$ \\
\bottomrule
\end{tabular}
}
\end{table}

\section{Related Work}
\label{sec:related}

\noindent\textbf{Relation Hallucination in MLLMs: }
MLLM hallucination research has often emphasized object existence, as in POPE~\cite{Li-hallucination-2023}.
Relation hallucination instead concerns image-unsupported relations among visible entities.
Reefknot~\cite{zheng2025reefknot} and Hal-Eval~\cite{jiang2024hal} provide relation-focused and fine-grained evaluations, while MMRA~\cite{wu2026mmra} extends relational association across granularities and images.
Beyond evaluation, CAP~\cite{wu2025mitigating} addresses spatial-relation errors through prompting constraints, while RVE~\cite{qin2026mitigating} enhances action-relevant regions.
DtC~\cite{zheng2025reefknot} combines confidence-based detection with intermediate-layer calibration to mitigate relation hallucination.

\noindent\textbf{Visual Evidence Calibration: }
Training-free methods enhance visual grounding through attention and token-level calibration~\cite{liu2024paying,yin2025clearsight,zhuang2025vasparse} or visual-evidence recovery~\cite{wang2024mllmseedynamiccorrection,zou2024look}.
OPERA~\cite{huang2024opera} links hallucination to attention aggregation onto generated summary tokens and penalizes over-trust in them.
Visual-sink analyses~\cite{kang2025see} identify highly attended tokens with limited semantic content, motivating attention redistribution to other visual tokens.
We examine how dominant visual evidence persists across autoregressive decoding steps, complementing studies of within-step attention concentration.
This persistence can involve informative tokens even when overall image attention remains strong.
Temporal approaches also detect attention defocus in text windows or guide attention with an early visual anchor~\cite{zhao2026attention,yao2026adapt}.
\ours uses an evolving token-specific reference to couple reinforcement of newly relevant evidence with attenuation conditioned on repeated dominance.
Appendix~\ref{sec:temporal_comparison} compares these signals and interventions.

\section{Conclusion}
\label{conclusion}
Relation hallucinations can occur despite correct object recognition. In the relation descriptions analyzed here, hallucinated responses exhibit more persistent visual focus than correct ones, reflecting a form of visual inertia.
Increasing previous-step attention carryover degrades relation prediction more than an entropy-matched current-step reference.
Motivated by these observations, we introduced Inertia-aware Visual Excitation (\ours), a lightweight, training-free decoding-time method that uses each visual token's recent attention history to reweight its contributions.
\ours favors newly relevant evidence while mildly attenuating persistently dominant contributions.
Across three MLLMs, \ours improves relation prediction on Reefknot and Hal-Eval.
These gains extend to the multi-image relational setting of MMRA, while performance on broader multimodal benchmarks is preserved.

\subsection*{AI use statement}
Generative AI tools assisted with literature review, experimental-design
discussions, code development and experiment execution, result analysis,
manuscript editing, and figure and table preparation. Research decisions
and interpretation remained under author direction. The authors take
responsibility for the final manuscript and all AI-assisted content,
including citations, code, experimental results, and scientific claims.

\subsection*{Ethics statement}
IVE is intended to improve the reliability of multimodal predictions
by reducing relation hallucinations. Our evaluation uses existing
multimodal benchmarks. Although IVE reduces hallucination rates in the
evaluated settings, the underlying models can still produce unsupported
statements and retain biases inherited from pretraining data.
Applications involving consequential decisions require task-specific
validation and appropriate human oversight.

\subsection*{Reproducibility statement}
Section~\ref{sec:approach} specifies the history-conditioned selection
and visual-value reweighting rules of IVE.
Appendix~\ref{sec:implementation} provides the decoding procedure and
model-specific selection layers, while Section~\ref{exp} describes
experimental settings and evaluation metrics.
The appendix additionally documents attention-displacement analyses,
carryover controls, and configuration studies to support reproduction
of the reported comparisons.

\bibliography{main}
\bibliographystyle{iclr2027_conference}

\appendix
\newpage

\setcounter{section}{0}
\setcounter{subsection}{0}
\renewcommand{\thesection}{\Alph{section}}
\renewcommand{\thesubsection}{\thesection.\arabic{subsection}}

\section*{Appendix Overview}
This appendix provides supporting attention analyses, implementation details, extended evaluations, and qualitative examples. The contents are organized as follows:
\begin{itemize}
    \setlength{\itemsep}{0pt}
    \item \textbf{Details for Motivation Analyses (\S\ref{sec:motivation_analyses}):} Attention-displacement analyses, cross-backbone distance comparisons, and carryover controls supporting Section~\ref{sec:pre}.
    \item \textbf{More Details of Implementation (\S\ref{sec:implementation}):} Configuration selection, within-forward value recalibration, detailed pseudocode, and comparison with temporal attention methods.
    \item \textbf{More Experimental Results and Analysis (\S\ref{sec:more_results_appendix}):} Type-wise Hal-Eval results, relevance-signal and random-control comparisons, POPE and CHAIR evaluations, broader capability checks, and configuration studies.
    \item \textbf{Case Studies (\S\ref{sec:case_studies_appendix}):} Qualitative comparisons between Vanilla and \ours.
\end{itemize}
\hrule
\vspace{4mm}

\section{Details for Motivation Analyses}
\label{sec:motivation_analyses}

\subsection{Attention Displacement Under Naive Amplification}
To further examine why naive visual attention amplification is insufficient for relation-hallucination mitigation, we compare the temporal attention patterns produced by the baseline model and the amplification method PAI~\cite{liu2024paying}.
Overall attention strength describes the influence assigned to the image. Comparing visual-attention distributions across output positions further reveals whether the same visual-token contributions remain dominant throughout a response.

Specifically, we generate 1,000 captions using LLaVA-1.5~\cite{liu2024improved} on a subset of COCO~\cite{lin2014microsoft}.
During generation, we store the normalized visual attention weights of the first 100 generated tokens.
For each sample, we then compute a pairwise distance matrix ($100 \times 100$) over visual-attention patterns using the same one-dimensional Wasserstein distance over flattened patch indices as in Equation~\ref{eq:pairwise_displacement}.
This matrix covers both nearby and distant output positions, providing a trajectory-level view of whether similar visual-attention patterns persist over an extended portion of the response.

Normalizing the visual-token slice separates its relative allocation from the total attention assigned to the image. The index-based ground cost preserves the fixed visual-token ordering and assigns larger costs to more widely separated indices. In the pairwise matrix, entries near the diagonal compare nearby output positions, while entries farther from it compare more widely separated stages of generation. Extended low-distance regions thus indicate that similar indexed attention distributions recur across an interval of the response.

\begin{figure}[t]
    \centering
    \includegraphics[width=.99\linewidth]{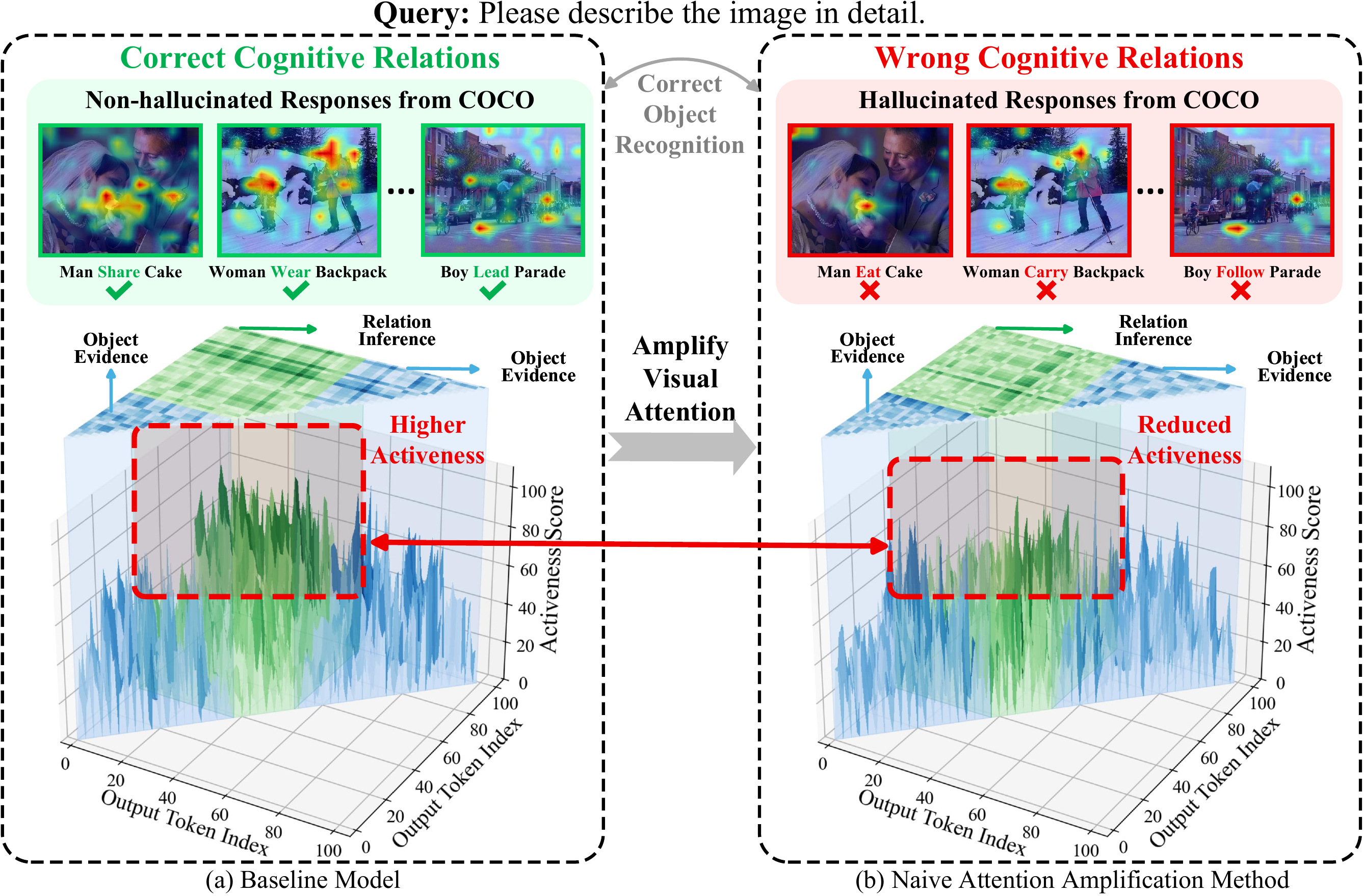}
    \caption{\textbf{Pairwise visual-attention displacement under naive amplification.} On the COCO captioning analysis, PAI exhibits lower pairwise displacement than the baseline across many output-token pairs. This supporting result concerns the effect of amplification and is distinct from the outcome-linked Reefknot comparison in the main text.}
    \label{fig:activeness_analysis}
\end{figure}

PAI exhibits lower displacement than Vanilla across many output-position pairs (Figure~\ref{fig:activeness_analysis}). Its normalized visual-attention distributions therefore remain more similar under the fixed token ordering over the course of generation. In this captioning comparison, amplification is accompanied by more persistent visual focus, extending the analysis beyond the overall attention strength measured in the main text.

\begin{figure}[t]
    \centering
    \includegraphics[width=.99\linewidth]{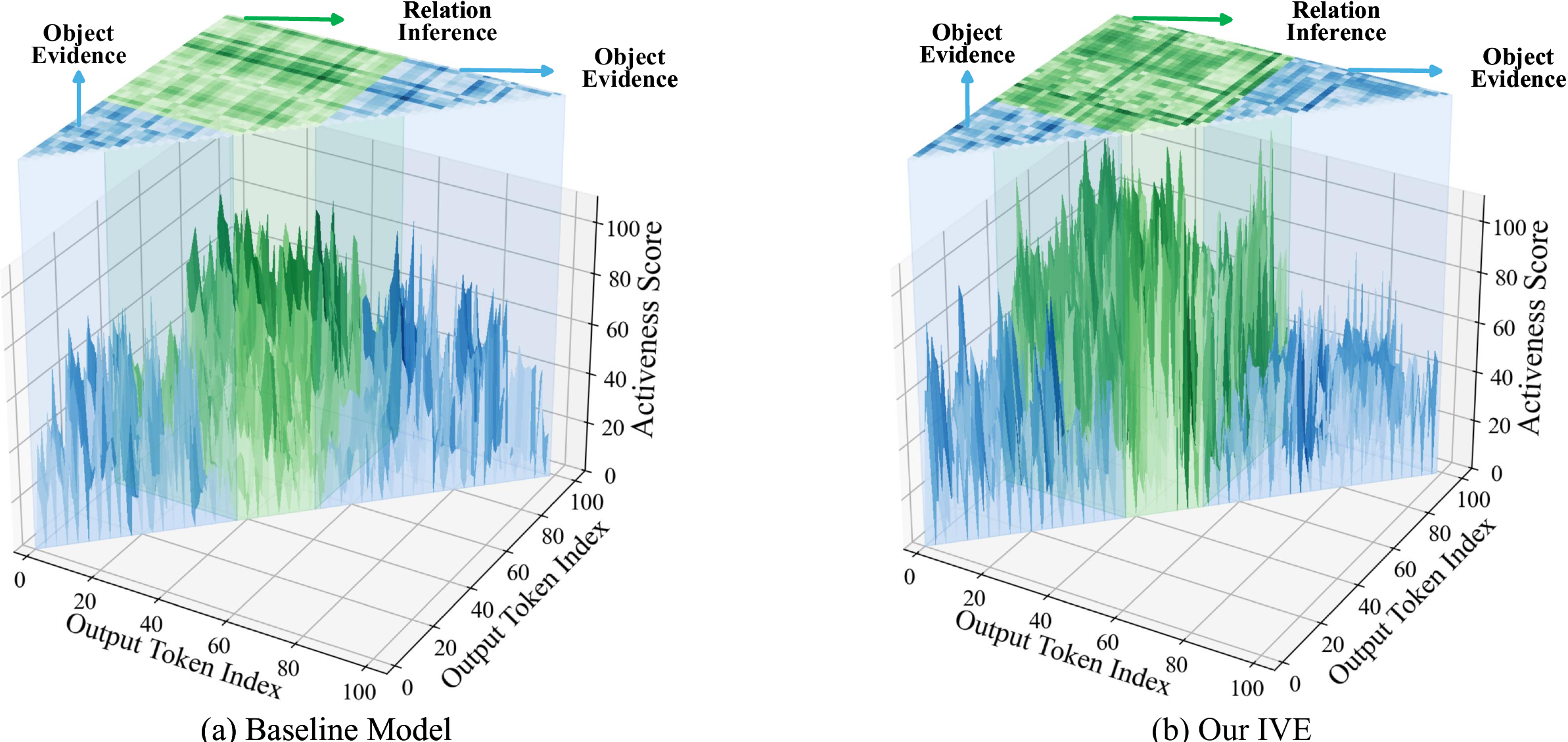}
    \caption{Pairwise visual-attention displacement for Vanilla and \ours on COCO captioning samples.}
    \label{fig:activeness_ours}
\end{figure}

Figure~\ref{fig:activeness_ours} compares Vanilla and \ours using the same normalization and flattened-index ground cost. Together, the two comparisons examine temporal attention patterns under both amplification and history-conditioned contribution reweighting. The pairwise view complements the consecutive-step average by retaining changes over short and long intervals.
These COCO analyses compare decoding conditions during captioning. Their role differs from the main-text Reefknot surfaces, which compare prediction outcomes under one Vanilla configuration. The former describe how attention trajectories vary with the intervention, whereas the latter connect persistent visual focus to relation-prediction outcomes.

\subsection{Relation-Description Protocol and Attention Displacement}
\label{sec:consecutive_displacement}
\label{sec:relation_description_protocol}

\paragraph{Task and generation.}
Each sample pairs an image with two entities specified in a Reefknot Y/N relation question. We generate a relation description with Vanilla LLaVA-1.5-7B using greedy decoding and a maximum of 128 new tokens. The task request is:
\begin{quote}
Please describe the relationship between the [entity A] and the [entity B] in the image.
\end{quote}
The entity placeholders are filled from the source question, and the request leaves response length unconstrained within the decoding limit. All candidate responses are generated under the same configuration before outcome grouping.

\paragraph{Outcome annotation.}
A target relation is \textit{correct} when the response explicitly describes an image-supported relation between the specified entities. It is \textit{unsupported} when the response makes an image-unsupported claim about that relation, including responses that mix supported and unsupported target-relation statements. Responses that omit the target relation or are too ambiguous to judge are labeled \textit{unanswered/ambiguous}. A partially informative response counts as correct only if it still expresses a supported target relation without a conflicting target-relation claim. Additional unsupported entity, attribute, or relation claims are recorded separately from target correctness.

\paragraph{Outcome groups.}
For the main-text surfaces, the correct group requires a correct target relation and no additional hallucinated claims. The hallucinated group contains responses with an unsupported target relation. Unanswered or ambiguous responses, and target-correct responses containing additional hallucinations, are excluded from this two-group comparison. After annotation, we retain responses with at least 50 valid output tokens and randomly select 500 eligible responses per group using seed 42. Selection is independent of the measured attention displacement. We compare separately sampled outcome groups, aligning attention trajectories by output-token index. The contrast characterizes a group-level association between attention persistence and relation errors.

\paragraph{Attention positions and aggregation.}
We record attention at the transformer block denoted by $l_s=20$, averaging over its heads and normalizing the visual-token slice as in Equation~\ref{eq:normalized_visual_attention}. Position $t$ refers to the query that predicts the corresponding output token. We use the first 50 valid output-token positions, excluding BOS, EOS, and padding tokens, and retain only responses that provide all 50 positions. Shorter responses are excluded without padding, repetition, or interpolation. The $24\times24$ patches are flattened in row-major order to indices $0,\ldots,575$. We compute the weighted one-dimensional Wasserstein distance with SciPy's \texttt{wasserstein\_distance}, passing the indices as support values and the normalized attention distributions as weights. The ground cost is $|a-b|$, giving distances in $[0,575]$. Each response contributes one $50\times50$ matrix, and we average these matrices elementwise within each outcome group, giving every selected response equal weight.

\paragraph{Independent prediction-level analysis.}
The consecutive-displacement analysis uses a separate Reefknot sample cohort with the same relation-description request, generation settings, and target-relation annotation rules, retaining its natural correctness distribution. Relation accuracy counts target-correct responses over all evaluated responses in each displacement group, including unanswered or ambiguous responses in the denominator. Target-relation correctness is assessed independently of additional hallucinations elsewhere in the response.
A pairwise matrix describes the temporal structure of one trajectory, but a single summary is useful when comparing many predictions with their outcomes. We summarize each trajectory by averaging Wasserstein distance over consecutive output positions:
\begin{equation}
    D_{\mathrm{attn}}=
    \frac{1}{T-1}\sum_{t=1}^{T-1}
    W_{\mathrm{idx}}\!\left(\mathbf{p}_t,\mathbf{p}_{t+1}\right),
    \label{eq:consecutive_displacement}
\end{equation}
where $T$ is the number of analyzed output positions and $\mathbf{p}_t$ is the normalized visual-attention distribution defined in Equation~\ref{eq:normalized_visual_attention}.
Each term measures the change from one output position to the next, and averaging prevents longer analyzed sequences from receiving a larger score simply because they contain more transitions. A low value indicates relatively similar attention distributions at successive positions. A higher value indicates greater mean change under the fixed flattened-token ordering.

\begin{table}[t]
    \centering
    \caption{Relation accuracy across consecutive attention-displacement groups on Reefknot.}
    \label{tab:displacement_groups}
    \footnotesize
    \begin{tabular*}{\linewidth}{@{\extracolsep{\fill}}lccc@{}}
        \toprule
        Group & $D_{\mathrm{attn}}$ range & Samples & Relation accuracy $\uparrow$ \\
        \midrule
        Low & $[0,32)$ & $\sim$330 & 58.3\% \\
        Medium & $[32,48)$ & $\sim$340 & 64.7\% \\
        High & $[48,\infty)$ & $\sim$330 & 71.2\% \\
        \bottomrule
    \end{tabular*}
\end{table}

Relation accuracy increases with consecutive attention displacement (Table~\ref{tab:displacement_groups}). The same statistic has a positive Spearman correlation with relation accuracy ($\rho=+0.31$) and a negative correlation with the corresponding hallucination rate ($\rho=-0.28$), with $p<0.001$ in both cases. These results associate changes in visual focus with prediction outcome on this separate sample.

The two summaries organize their samples differently. The main-text surfaces first separate predictions by correctness and then compare their trajectories. Here, predictions are grouped by displacement, and relation accuracy is measured within each group. Each displacement group contains both correct and incorrect predictions. The increasing accuracy across groups provides complementary evidence for the association between attention change and relation accuracy.

\subsection{Attention Displacement Across Backbones}
\label{sec:distance_robustness}

We compare Vanilla, PAI, and \ours on Reefknot to examine whether their attention-displacement patterns are consistent across LLaVA-1.5 and InstructBLIP. For InstructBLIP, we project its LLM-side Q-Former query attention through Q-Former cross-attention to obtain a patch-level distribution, flatten the resulting patch order, and then compute the same index-based Wasserstein distance. Each visual encoder defines its own support and patch indexing, so we compare method rankings within each backbone.

We also report Jensen--Shannon distance, denoted $D_{\mathrm{JS}}$, which is invariant to applying the same permutation of patch indices to both distributions and measures allocation changes independently of index separation. Together, the Wasserstein and Jensen--Shannon comparisons test whether the within-backbone method ordering persists across distance measures.

\begin{table}[t]
\centering
\caption{Attention distances and Y/N $R_{score}$ for LLaVA-1.5 and InstructBLIP. $D_{\mathrm{attn}}$ uses Wasserstein distance over flattened patch indices, and $D_{\mathrm{JS}}$ uses Jensen--Shannon distance.}
\label{tab:distance_robustness}
\footnotesize
\begin{tabularx}{\linewidth}{l*{3}{>{\centering\arraybackslash}X}}
\toprule
\multicolumn{4}{c}{\textbf{(a) LLaVA-1.5}} \\
\midrule
\textbf{Method} & $D_{\mathrm{attn}}$ & $D_{\mathrm{JS}}$ & $R_{score}\uparrow$ \\
\midrule
Vanilla & 41.52 & 0.193 & 64.39 \\
+PAI & 38.64 & 0.181 & 64.71 \\
\rowcolor[HTML]{E8F2FE}
\textbf{+\ours} & \textbf{47.21} & \textbf{0.224} & \textbf{67.21} \\
\bottomrule
\end{tabularx}

\vspace{1.5mm}
\begin{tabularx}{\linewidth}{l*{3}{>{\centering\arraybackslash}X}}
\toprule
\multicolumn{4}{c}{\textbf{(b) InstructBLIP}} \\
\midrule
\textbf{Method} & $D_{\mathrm{attn}}$ & $D_{\mathrm{JS}}$ & $R_{score}\uparrow$ \\
\midrule
Vanilla & 39.83 & 0.187 & 56.28 \\
+PAI & 37.12 & 0.174 & 56.80 \\
\rowcolor[HTML]{E8F2FE}
\textbf{+\ours} & \textbf{45.67} & \textbf{0.216} & \textbf{59.09} \\
\bottomrule
\end{tabularx}
\end{table}

Both distance measures give the same ordering for both backbones, with lower attention change under PAI than Vanilla and higher change under \ours (Table~\ref{tab:distance_robustness}). The agreement across backbones and distance choices supports the consistency of this temporal pattern.

PAI modestly improves $R_{score}$ despite reducing the two attention-distance measures, while \ours combines larger attention changes with the strongest performance in each backbone.

\subsection{Progressive Carryover Control}
\label{sec:carryover_control}

The carryover intervention in Equation~\ref{eq:attention_carryover} mixes the previous-step visual-attention distribution into the current one. Increasing the mixing weight makes the previous spatial pattern more influential, while the current distribution continues to contribute. To compare temporal carryover at a similar level of attention concentration, the matched control uses the same entropy but reshapes only the current-step distribution:
\begin{equation}
    \mathbf{p}^{\mathrm{ctrl}}_t(\lambda)
    =\operatorname{Norm}\!\left(\mathbf{p}_t^{\,\beta_t}\right),
    \qquad
    \mathcal{H}\!\left(\mathbf{p}^{\mathrm{ctrl}}_t\right)
    =\mathcal{H}\!\left(\mathbf{p}^{\mathrm{carry}}_t\right).
    \label{eq:entropy_matched_control}
\end{equation}
At each step, $\beta_t>0$ is chosen so that the control and carryover distributions have the same entropy. Here, $\operatorname{Norm}$ denotes $\ell_1$ normalization and $\mathcal{H}$ denotes entropy. The resulting relative distribution is mapped back to the visual-token attention slice while preserving its total mass.

We initialize the previous-attention state from the last eligible input-text query before the final prompt position. Carryover and its matched reference are applied before attention--value aggregation, with outputs selected from the resulting logits. The previous-attention state advances once per query.

The exponent adjusts concentration while preserving the ordering of the current attention weights. Values above one sharpen the distribution toward its dominant entries, whereas values below one flatten it. This current-step transformation contains no previous-step spatial allocation, and preserving total visual-attention mass isolates redistribution within the visual slice.

Under progressive carryover, both $R_{score}$ and consecutive displacement decline more strongly than under the matched reference. The widening performance gap in Table~\ref{tab:progressive_carryover} supports a contribution from temporal persistence under this intervention. Entropy specifies the concentration being matched, while displacement records the accompanying change in the trajectory.

\paragraph{Recovery under induced carryover.}
We additionally apply \ours under carryover at $\lambda=0.5$ to examine whether history-conditioned reweighting remains useful when persistence is explicitly strengthened. The relevant comparison is between carryover alone and carryover with \ours, with Vanilla providing the no-carryover reference. The combined condition follows the initialization and value-update procedure in Appendix~\ref{sec:implementation}. \ours partially recovers the lost $R_{score}$ and moves both attention displacement and the $Cogn.$ hallucination rate toward their Vanilla levels (Table~\ref{tab:carryover_rescue}). The joint improvement connects the method's behavioral benefit with a change in the induced attention pattern.

\begin{table}[t]
    \centering
    \caption{IVE under induced previous-step carryover at $\lambda=0.5$ on Reefknot Y/N.}
    \label{tab:carryover_rescue}
    \footnotesize
    \begin{tabular*}{\linewidth}{@{\extracolsep{\fill}}lccc@{}}
        \toprule
        Condition & $D_{\mathrm{attn}}$ & Cogn.$\downarrow$ & $R_{score}\uparrow$ \\
        \midrule
        Vanilla & 41.52 & 33.99 & 64.39 \\
        Carryover & 28.73 & 39.84 & 54.50 \\
        Carryover + IVE & 36.81 & 35.12 & 59.77 \\
        \bottomrule
    \end{tabular*}
\end{table}

\subsection{Manual Intervention Settings}
\label{sec:manual_interventions}
For the intervention analysis in Table~\ref{tab:intervention}, we apply several manual perturbations to the visual attention maps of LLaVA-1.5~\cite{liu2024improved} on Reefknot~\cite{zheng2025reefknot}.
These settings provide a broader comparison of attention perturbations alongside the progressive carryover analysis. All perturbations precede attention--value aggregation, with input-initialized attention history for the temporal conditions (Appendix~\ref{sec:carryover_control}).
\begin{table}[t]
    \centering
    \caption{Supporting attention interventions on Reefknot Y/N. Previous-step propagation corresponds to the $\lambda=1$ carryover condition in Table~\ref{tab:progressive_carryover}. $D_{\mathrm{attn}}$ is mean consecutive attention displacement.}
    \label{tab:intervention}
    \footnotesize
    \begin{tabular*}{\linewidth}{@{\extracolsep{\fill}}lcc@{}}
        \toprule
        \textbf{Intervention} & $D_{\mathrm{attn}}$ & $R_{score}\uparrow$ \\
        \midrule
        Vanilla & 41.52 & 64.39 \\
        Random perturbation & 38.76 & 62.27 \\
        Shuffled previous-step & 34.21 & 59.36 \\
        Current-step focus & 30.47 & 56.61 \\
        Previous-step propagation & 19.24 & 47.68 \\
        \bottomrule
    \end{tabular*}
\end{table}

\begin{itemize}
    \item \textbf{Random perturbation.} We add random noise to the current-step visual attention map and then re-normalize it, introducing a generic disturbance without carrying forward the previous spatial pattern.
    \item \textbf{Shuffled previous-step attention.} We inject previous-step attention after shuffling its spatial positions. This retains the previous attention values while changing which regions receive them.
    \item \textbf{Current-step focus.} We sharpen the current-step distribution toward its already dominant regions, increasing concentration without introducing previous-step attention.
    \item \textbf{Previous-step propagation.} We mix the current- and previous-step distributions with equal weight and normalize the result, corresponding to $\lambda=1$ in Equation~\ref{eq:attention_carryover}.
\end{itemize}
The shuffled and unshuffled conditions test whether previous attention values retain their spatial assignment. Current-step focus strengthens regions that already dominate the present distribution, and random perturbation supplies a generic disturbance. These conditions compare distinct perturbation schemes at their respective intervention strengths.

Previous-step propagation produces the largest $R_{score}$ reduction among these settings, accompanied by the lowest consecutive displacement. Its $\lambda=1$ setting combines the current and previous distributions with equal weight. The entropy-matched comparison in Section~\ref{sec:pre} separately examines carryover at comparable attention concentration.

\section{More Details of Implementation}
\label{sec:implementation}

Token selection uses head-averaged attention over all tokens at layer 20 for LLaVA-1.5~\cite{liu2024improved}, layer 20 for InstructBLIP~\cite{dai2023instructblip}, and layer 18 for Qwen2.5-VL-Instruct~\cite{bai2025qwen2}. Value recalibration spans all attention heads and transformer layers. Displacement analysis uses the unit-normalized visual slice at output positions.

\paragraph{Reefknot evaluation score.}
For each Reefknot task $t$, we report its task-conditioned component of the benchmark score:
\[
    R_{score}(t)=100-
    \frac{N_{P,t}H_{P,t}+N_{C,t}H_{C,t}}{N_{P,t}+N_{C,t}}.
\]
Here, $H_{P,t}$ and $H_{C,t}$ are the perception- and cognition-hallucination rates in percent, and $N_{P,t}$ and $N_{C,t}$ are the corresponding sample counts from the released splits. We report this score separately for Y/N and MCQ.

\paragraph{Hal-Eval metrics.}
For the discriminative evaluations in Tables~\ref{tab:haleval} and~\ref{tab:haleval_types}, the positive class is an image-consistent description (\textit{Yes}). Accuracy and binary F1 are computed from the same predictions. Within each reported evaluation condition, we pool the confusion counts over examples and compute $F_1=2TP/(2TP+FP+FN)$. Both metrics are reported as percentages.

\paragraph{Decoding settings.}
Table~\ref{tab:decoding_robustness} compares greedy decoding, beam search with four beams, and nucleus sampling with top-$p=0.9$, temperature 1.0, and no top-$k$ truncation. For each backbone and method, sampling uses five runs with seeds 0, 1, 2, 3, and 4. We report the mean and sample standard deviation of the five full-evaluation $R_{score}$ values, using the same decoding settings across methods.

\paragraph{Configuration selection.}
We select $\alpha$, $\gamma$, $\tau$, and the token-selection layer separately for each backbone using only held-out Reefknot validation data, disjoint from the final evaluation samples. Each configuration is fixed before testing and reused across that backbone's benchmark evaluations, including Hal-Eval, POPE, and MMRA. The unified configuration is also fixed before evaluation and shared across the three backbones. The coefficient and layer sweeps in Tables~\ref{tab:coefficient} and~\ref{tab:layer_ablation} examine sensitivity around the selected configurations.

\paragraph{Within-forward value recalibration.}
We retain fixed copies of the original visual values separately from the working cache. Before generation, the EMA reference, inertia counts, and history length are accumulated in causal order from image-attending input-text queries preceding the final prompt position, excluding padding and visual-token queries, while visual values remain unchanged. For each output-producing query, the hook at $l_s$ reads attention before attention--value aggregation and computes token sets and factors from the preceding history. Emergent tokens are selected using the original relevance scores, and their multipliers are capped at $1.10$ before cache updates. At the reported thresholds, $1+S_{t,j}>1.10$ for every selected emergent token, so its applied multiplier is $1.10$.

At each query, every visual-token entry in the working cache is assigned its capped factor times the corresponding original value, across all layers and heads. Tokens outside both selected sets have factor one and are restored to their original values. The original copies remain unchanged, and the active tensor at $l_s$ uses the reconstructed entries without reapplying the factors.

Layer $l_s$ and later layers use the adjusted values in the current forward pass. Earlier layers have already executed and use the newly assigned cache entries on the next query. Each output is produced by one forward pass, with its resulting logits passed to the original decoding rule.

\paragraph{History updates and persistence counts.}
After value recalibration, the pre-intervention attention observation updates the EMA with weights $\gamma$ and $1-\gamma$ for the old reference and new observation. The reported coefficients give greater weight to the latest observation. Inertia counts and history length advance once, excluding the current query from its own factor. A query without history uses identity modulation and initializes the reference.

The frequency $C_{t,j}/n_t$ determines the inertia factor within $[1-\alpha,1]$. With the reinforcement cap, the applied value multiplier lies in $[1-\alpha,1.10]$ relative to the fixed original value at every query. The EMA, inertia counts, and history length evolve across queries, while each value update is computed from the original value.

State is local to each sample and beam hypothesis, following cache branching and reordering. Algorithm~\ref{alg:ive} uses $(\mathbf{b},\mathbf{C},n)$ for the preceding EMA, inertia counts, and history length. Reported runtimes include input processing and cache updates.

\begin{algorithm}[t]
\caption{Within-forward recalibration from fixed original values in \ours.}
\label{alg:ive}
\footnotesize
\DontPrintSemicolon
\SetAlgoVlined
\KwIn{MLLM $f$ with $L$ layers, final input token $x_P$, KV cache for $\mathbf{x}_{<P}$, fixed original visual values $\{\mathbf{V}^{(l,h),\mathrm{base}}_j\}$, history state $(\mathbf{b},\mathbf{C},n)$, visual-token set $\mathcal{V}$, selection layer $l_s$, maximum output length $T_{\max}$, threshold $\tau$, coefficients $\alpha,\gamma$}
\KwOut{Generated response}
$\epsilon\gets10^{-6}$, $\mathbf{y}\gets\varnothing$\;
$q\gets x_P$ with its original position and causal mask\;
\For{$t\gets1$ \KwTo $T_{\max}$}{
    Process $q$ through layers $1,\ldots,l_s-1$, extending the cache\;
    Compute $\{\mathbf{A}^{(h)}_t\}_{h=1}^{N_h}$ at layer $l_s$ before value aggregation\;
    $a_{t,j}\gets N_h^{-1}\sum_{h=1}^{N_h}\mathbf{A}^{(h)}_{t,j}$
    for every $j\in\mathcal{V}$\;
    \eIf{$n>0$}{
        Compute $\mathbf{S}_t,\mathcal{E}_t,\mathcal{I}_t$ from $\mathbf{a}_t,\mathbf{b}$ using Equations~\ref{eq:history_relevance} and~\ref{eq:visual_token_sets}\;
        Compute $m_{t,j}$ using Equation~\ref{eq:modulation_factor} with $C_{t,j}=C_j$ and $n_t=n$\;
        Cap $m_{t,j}$ at $1.10$ for $j\in\mathcal{E}_t$\;
    }{
        $\mathcal{E}_t\gets\varnothing$, $\mathcal{I}_t\gets\varnothing$, and $m_{t,j}\gets1$ for $j\in\mathcal{V}$\;
    }
    \ForEach{transformer layer $l$ and attention head $h$}{
        $\mathbf{V}^{(l,h)}_j\gets
        m_{t,j}\mathbf{V}^{(l,h),\mathrm{base}}_j$ for every visual token $j\in\mathcal{V}$\;
    }
    Use the updated visual entries in layer $l_s$'s active value tensor\;
    Complete layer $l_s$ and layers $l_s+1,\ldots,L$, without reapplying the factors\;
    Compute $\mathbf{z}_t$ from the resulting final hidden state\;
    Select $y_t$ from $\mathbf{z}_t$ with the base decoding rule and append it to $\mathbf{y}$\;
    \eIf{$n=0$}{
        $\mathbf{b}\gets\mathbf{a}_t$\;
    }{
        $\mathbf{b}\gets\gamma\mathbf{b}+(1-\gamma)\mathbf{a}_t$\;
    }
    $C_j\gets C_j+\mathbf{1}[j\in\mathcal{I}_t]$ for $j\in\mathcal{V}$, and $n\gets n+1$\;
    \If{$y_t$ is the end-of-sequence token}{
        \textbf{break}\;
    }
    $q\gets y_t$ at the next sequence position\;
}
\Return{$\mathbf{y}$}\;
\end{algorithm}

\begin{figure}[t]
    \centering
    \includegraphics[width=\linewidth]{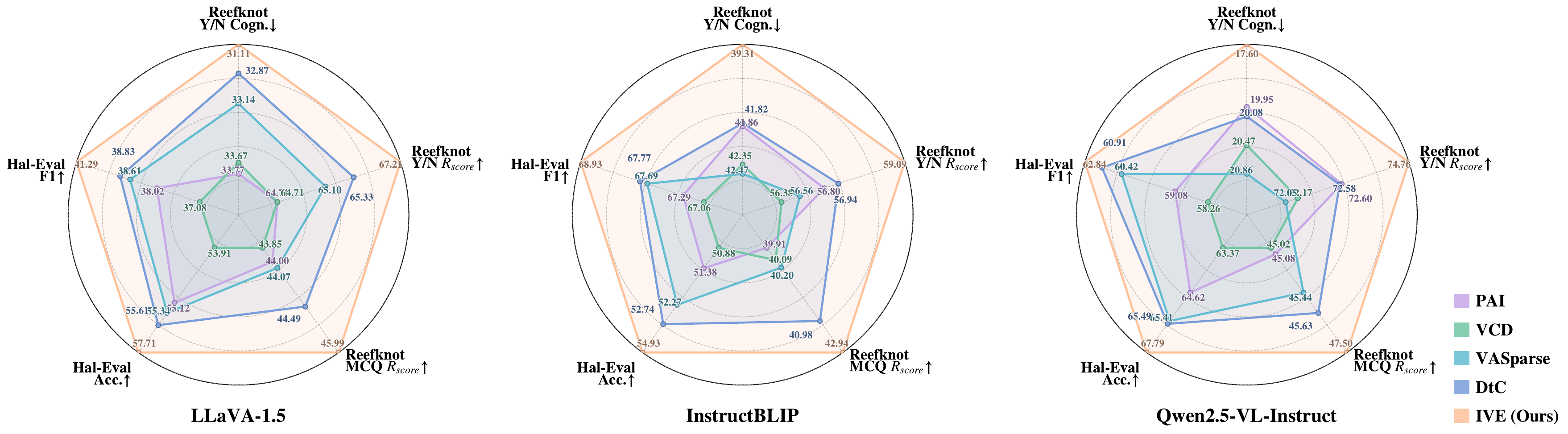}
    \caption{Radar comparison on Reefknot and Hal-Eval across three MLLMs. Each axis uses a padded raw-score range whose favorable endpoint is the best reported value. The Y/N Cogn. axis is reversed so that farther outward consistently indicates better performance.}
    \label{fig:radar}
\end{figure}

\begin{figure}[t]
    \centering
    \includegraphics[width=\linewidth]{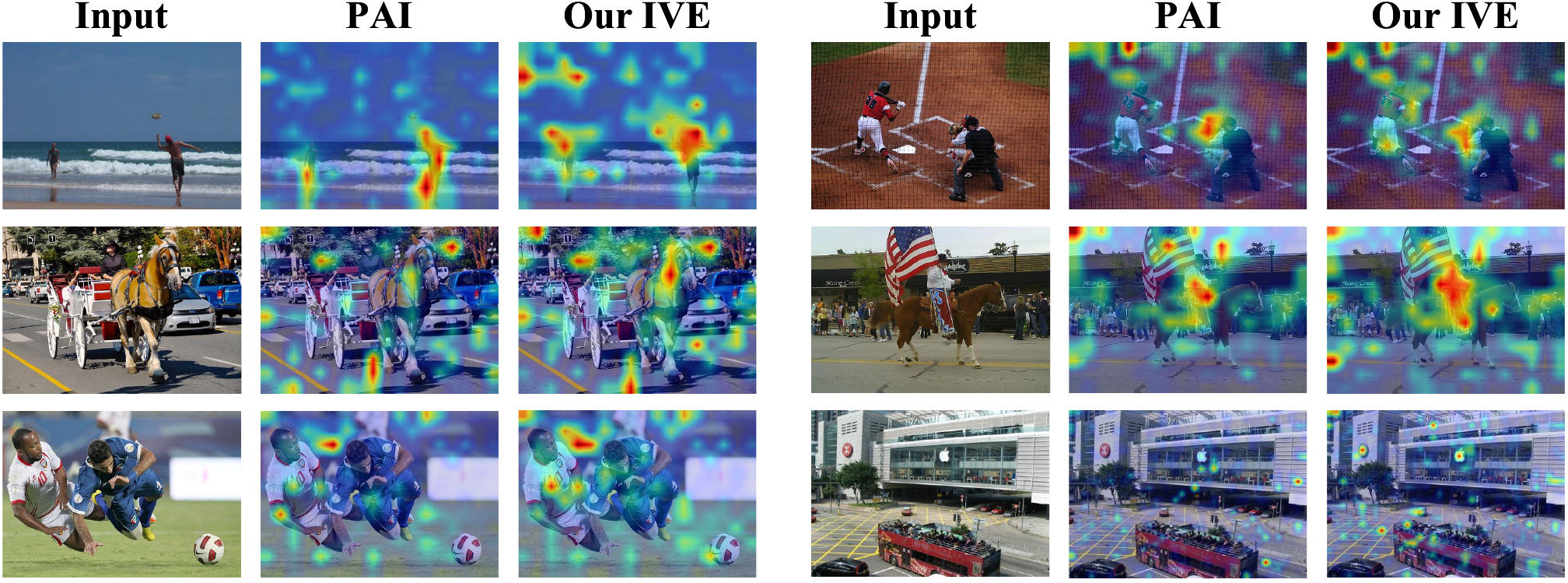}
    \caption{Visual-attention heatmaps for \ours and the amplification method PAI.}
    \label{fig:heatmap}
\end{figure}

\subsection{Comparison with Temporal Attention Methods}
\label{sec:temporal_comparison}

Temporal attention provides information about both the changing relevance of visual tokens and the deterioration of visual grounding during generation. The methods below use this information at different levels, from individual visual contributions to candidate text continuations. Table~\ref{tab:temporal_method_comparison} summarizes their reference signals and interventions. Our analysis focuses on persistence associated with relation errors and examines the effect of increased previous-step carryover at comparable attention concentration.

\begin{table}[t]
\centering
\caption{Reference signals and interventions in temporal attention methods.}
\label{tab:temporal_method_comparison}
\footnotesize
\begin{tabular*}{\linewidth}{@{\extracolsep{\fill}}>{\raggedright\arraybackslash}p{0.18\linewidth}>{\raggedright\arraybackslash}p{0.32\linewidth}>{\raggedright\arraybackslash}p{0.42\linewidth}@{}}
\toprule
\textbf{Method} & \textbf{Reference or signal} & \textbf{Intervention} \\
\midrule
Attention Never Lie & Learned detection from windowed attention features & Detector-guided selection of candidate continuations \\
\addlinespace
ADAPT & Attention degradation relative to an early visual anchor & Sparse anchor-guided attention steering, with preference tuning in the full method \\
\addlinespace
\ours & EMA-relative attention rise and prior inertia-selection frequency & Paired reinforcement and conditional attenuation of visual values \\
\bottomrule
\end{tabular*}
\end{table}

\paragraph{Attention Never Lie.}
Attention Never Lie~\cite{zhao2026attention} links hallucination to visual-attention defocus and examines attention differences between nearby and more distant output positions. Its correction mechanism trains a detector on attention features from text windows and uses detector scores to select candidate continuations. Attention history guides continuation selection, and the paper also evaluates combining its detector with visual-attention amplification. \ours applies token-specific attention history directly to visual-value reinforcement and attenuation within the base decoding procedure.

\paragraph{ADAPT.}
ADAPT~\cite{yao2026adapt} characterizes attention degradation through fading visual focus, temporal rigidity, and spatial bias. It constructs a query-dependent visual anchor from early attention across multiple layers and uses degradation scores to trigger sparse, anchor-guided attention steering. This correction supports a training-free variant, while the full method additionally incorporates preference tuning. The early anchor guides attention toward stable visual evidence. \ours maintains an evolving reference for each token and conditions visual-value reinforcement and attenuation on its recent trajectory and inertia-selection history.

\paragraph{Visual Inertia terminology.}
GThinker~\cite{zhan2026gthinker} uses Visual Inertia to describe a model's tendency to retain its initial visual interpretation. Its cue-guided rethinking framework learns to revisit visual cues through supervised fine-tuning and reinforcement learning. In this paper, visual inertia denotes persistent reuse of previously dominant visual evidence across decoding, characterized through attention trajectories and carryover interventions.

The relevance-signal comparison in Table~\ref{tab:relevance_signal_comparison} evaluates token-scoring statistics under matched selection budgets and fixed reinforcement and attenuation factors. This diagnostic examines the selection signal separately from the full method's thresholding and frequency-dependent attenuation.

\section{More Experimental Results and Analysis}
\label{sec:more_results_appendix}

We extend the main evaluations with type-wise hallucination results, controlled comparisons of visual reweighting, and object-hallucination tests on both binary questions and generated captions. Additional analyses examine broader multimodal performance and the effects of coefficient and layer choices.

\subsection{Temporal-Attention Baselines on Reefknot}
\label{sec:temporal_baselines}

Table~\ref{tab:temporal_method_results} lists Vanilla, the training-free variant of ADAPT~\cite{yao2026adapt}, Attention Never Lie~\cite{zhao2026attention}, and \ours for comparison on Reefknot across the three evaluated backbones. We reproduce ADAPT-TF and Attention Never Lie following their official repositories and default inference protocols, without tuning on the Reefknot test set. ADAPT excludes preference-tuned components, while Attention Never Lie follows the repository-provided detector setup. All methods use matched backbones, prompts, decoding settings, and evaluation splits.

\begin{table}[t]
\centering
\caption{Comparison with temporal-attention methods on Reefknot across three backbones.}
\label{tab:temporal_method_results}
\footnotesize
\begin{tabular*}{\linewidth}{@{\extracolsep{\fill}}llcccc@{}}
\toprule
\multirow{2}{*}{\textbf{Backbone}} &
\multirow{2}{*}{\textbf{Method}} &
\multicolumn{2}{c}{\textbf{Y/N}} &
\multicolumn{2}{c}{\textbf{MCQ}} \\
\cmidrule(lr){3-4}\cmidrule(lr){5-6}
& & \textbf{Cogn.}$\downarrow$ & $R_{score}\uparrow$
& \textbf{Cogn.}$\downarrow$ & $R_{score}\uparrow$ \\
\midrule
\multirow{4}{*}{LLaVA-1.5}
& Vanilla & 33.99 & 64.39 & 51.04 & 43.70 \\
& ADAPT (training-free) & 32.87 & 65.34 & 50.31 & 44.72 \\
& Attention Never Lie$^{\dagger}$ & 32.36 & 65.90 & 49.86 & 45.35 \\
& \ours & \textbf{31.11} & \textbf{67.21} & \textbf{49.08} & \textbf{45.99} \\
\midrule
\multirow{4}{*}{InstructBLIP}
& Vanilla & 42.47 & 56.28 & 53.85 & 39.94 \\
& ADAPT (training-free) & 40.94 & 57.65 & 52.18 & 41.68 \\
& Attention Never Lie$^{\dagger}$ & 41.12 & 57.52 & 51.91 & 41.92 \\
& \ours & \textbf{39.31} & \textbf{59.09} & \textbf{50.87} & \textbf{42.94} \\
\midrule
\multirow{4}{*}{Qwen2.5-VL}
& Vanilla & 20.58 & 72.07 & 49.69 & 44.91 \\
& ADAPT (training-free) & 19.22 & 73.31 & 48.03 & 46.72 \\
& Attention Never Lie$^{\dagger}$ & 18.96 & 73.53 & 48.27 & 46.51 \\
& \ours & \textbf{17.60} & \textbf{74.76} & \textbf{47.01} & \textbf{47.50} \\
\bottomrule
\end{tabular*}
\par\smallskip
\begin{minipage}{\linewidth}
\footnotesize
$^{\dagger}$Uses a trained detector.
\end{minipage}
\end{table}

\ours achieves higher $R_{score}$ and lower $Cogn.$ than both temporal baselines on all three backbones in both Y/N and MCQ. The $R_{score}$ margin over the better-performing temporal baseline in each setting ranges from 0.64 to 1.44 points.

\subsection{Hal-Eval Results Across Hallucination Types}
\label{sec:haleval_types}

Hal-Eval~\cite{jiang2024hal} separates hallucinations into Object, Attribute, Relation, and Event categories.
Table~\ref{tab:haleval_types} complements the three-backbone Relation results in the main text by reporting all four categories for LLaVA-1.5.
This comparison asks whether the gains are larger for relations than for other hallucination types. \ours improves accuracy and F1 over both Vanilla and MemVR in all four categories, with its largest gains over Vanilla occurring on Relation. The type-wise pattern is consistent with the relation-focused motivation while retaining improvements on Object, Attribute, and Event.

Relation has the lowest Vanilla accuracy and F1 among the four categories. \ours narrows its performance gap to the other categories, although Relation remains the most difficult category on both metrics.

\begin{table}[t]
\centering
\caption{Accuracy and binary F1 (\%) across Hal-Eval hallucination types with LLaVA-1.5.}
\label{tab:haleval_types}
\footnotesize
\begin{tabular*}{\linewidth}{@{\extracolsep{\fill}}lcccccc@{}}
\toprule
\multirow{2}{*}{\textbf{Type}} &
\multicolumn{2}{c}{\textbf{Vanilla}} &
\multicolumn{2}{c}{\textbf{+MemVR}} &
\multicolumn{2}{c}{\textbf{+IVE (Ours)}} \\
\cmidrule(lr){2-3}\cmidrule(lr){4-5}\cmidrule(lr){6-7}
& \textbf{Acc.}$\uparrow$ & \textbf{F1}$\uparrow$
& \textbf{Acc.}$\uparrow$ & \textbf{F1}$\uparrow$
& \textbf{Acc.}$\uparrow$ & \textbf{F1}$\uparrow$ \\
\midrule
Object    & 62.20 & 48.50 & 63.70 & 50.10 & \textbf{64.10} & \textbf{50.80} \\
Attribute & 62.20 & 48.50 & 63.50 & 49.80 & \textbf{63.90} & \textbf{50.30} \\
Relation  & 52.65 & 35.82 & 55.83 & 39.04 & \textbf{57.71} & \textbf{41.29} \\
Event     & 62.70 & 58.90 & 64.10 & 60.40 & \textbf{64.80} & \textbf{61.20} \\
\bottomrule
\end{tabular*}
\end{table}

\subsection{Component and Selection-Signal Ablations}
\label{sec:component_ablation}

\paragraph{Reinforcement and attenuation.}
Table~\ref{tab:ablation_efficiency}(a) examines the two reweighting branches on LLaVA-1.5 using $l_s=20$, $\alpha=0.10$, $\gamma=0.09$, and $\tau=3.00$. Both single-branch conditions retain the full token-selection and history-update rules. Reinforce only sets the inertia-token multiplier to one, while Attenuate only sets the emergent-token multiplier to one. The enabled branch uses Equation~\ref{eq:modulation_factor}, including the $1.10$ reinforcement cap. Inertia counts continue to update when attenuation is disabled, and each condition maintains history along its own generated trajectory. Enabling both branches gives full \ours, while disabling both gives identity modulation. All conditions share evaluation samples, prompts, scoring, and deterministic decoding, with visual values reconstructed from fixed originals at each query.

\paragraph{Matched selection signals.}
To compare which tokens different signals prioritize, Table~\ref{tab:relevance_signal_comparison} uses a fixed-budget diagnostic on Reefknot Y/N with LLaVA-1.5. We average the emergent- and inertia-set sizes of full \ours over valid intervention queries on held-out validation data and round each mean to the nearest integer. This gives reinforcement and attenuation budgets of $K_+=8$ and $K_-=32$, fixed before testing. At each query, score-based conditions reinforce the highest-scoring $K_+$ visual tokens and attenuate the lowest-scoring $K_-$ among the remainder. Their multipliers are fixed at $1.10$ and $0.90$, respectively, relative to the original visual values, with all other entries restored to their originals. Evaluation inputs, decoding, and selection layer remain matched across signals. The IVE-score row uses this ranking and scaling protocol in place of the full method's threshold-based sets and frequency-dependent attenuation.

We rank unrounded scores and break exact ties by ascending original visual-token index in both selections. Random refresh uses an independent generator for each sample, initialized with seed $42+i$, where $i$ is the fixed sample index. Each query draws a fresh visual-token permutation, assigning the first $K_+$ entries to reinforcement and the next $K_-$ to attenuation. The generator advances across queries without reseeding, and the two selected sets remain disjoint.

\paragraph{Token-level scores.}
Cumulative attention measures past prominence, while previous-step differences retain the direction of the latest change. The IVE score in Equation~\ref{eq:history_relevance} combines a short-term EMA reference with attention-scale normalization. For entropy and concentration, we use each token's contribution to the corresponding statistic. With $p_{t,j}=a_{t,j}/\sum_{k\in\mathcal V}a_{t,k}$, these scores are
\[
    s^{\mathrm{ent}}_{t,j}=-p_{t,j}\log(p_{t,j}+\epsilon),
    \qquad
    s^{\mathrm{conc}}_{t,j}=p_{t,j}^{2},
    \qquad \epsilon=10^{-6}.
\]
Both provide token-level rankings from the current visual-attention distribution. The concentration score preserves the ordering of $p_{t,j}$ and therefore implements current-saliency selection under the matched budget.

The attention-sinking control uses the sink-dimension activation metric of~\citet{kang2025see}. Let $\mathbf h_j\in\mathbb R^D$ be visual token $j$'s prefill hidden state at the input to layer $l_s=20$. For LLaVA-1.5-7B, the sink dimensions are $\mathcal D_{\mathrm{sink}}=\{1415,2533\}$, giving
\[
    \phi_j=
    \frac{\max_{d\in\mathcal D_{\mathrm{sink}}}|\mathbf h_j[d]|}
    {\sqrt{D^{-1}\sum_{d=1}^{D}\mathbf h_j[d]^2}},
    \qquad s^{\mathrm{sink}}_{t,j}=-\phi_j.
\]
The negative sign assigns higher attenuation priority to tokens with larger sink activations. These scores remain fixed across output queries and use the same budgets and value multipliers as the other ranking controls.

\begin{table}[t]
\centering
\caption{Selection-signal comparison on Reefknot Y/N with LLaVA-1.5. All non-Vanilla rows use matched token budgets and fixed $1.10/0.90$ reinforcement/attenuation factors. $D_{\mathrm{attn}}$ measures consecutive attention displacement.}
\label{tab:relevance_signal_comparison}
\resizebox{\linewidth}{!}{%
\begin{tabular}{llcccc}
\toprule
\textbf{Signal} & \textbf{Property} & \textbf{Perc.}$\downarrow$ & \textbf{Cogn.}$\downarrow$ & $\mathbf{R_{score}}\uparrow$ & $D_{\mathrm{attn}}$ \\
\midrule
Vanilla & -- & 37.67 & 33.99 & 64.39 & 41.52 \\
Random refresh & No structure & 37.44 & 33.71 & 64.64 & 42.18 \\
Cumulative attention & Token-level, nondirectional & 37.12 & 33.28 & 65.02 & 43.51 \\
Previous-step difference & Token-level, directional & 36.89 & 32.97 & 65.30 & 44.23 \\
Entropy contribution & Token-level, current-only & 37.31 & 33.44 & 64.85 & 42.87 \\
HHI contribution & Token-level, current-only & 37.38 & 33.51 & 64.78 & 42.63 \\
Attention-sinking score & Token-level, nondirectional & 37.08 & 33.19 & 65.09 & 43.74 \\
\rowcolor[HTML]{E8F2FE}
\textbf{IVE score (matched budget)} & \textbf{Token-level, directional, smoothed} & \textbf{35.35} & \textbf{31.73} & \textbf{66.67} & \textbf{44.72} \\
\bottomrule
\end{tabular}
}
\end{table}

All tested signals improve $R_{score}$ over Vanilla, with previous-step difference performing best among the alternatives. The IVE score achieves the highest $R_{score}$ and the lowest hallucination rates in both reported categories under matched selection budgets and scaling factors. This comparison evaluates the ranking signal under fixed selection budgets and modulation strengths.

The history-conditioned score also produces the largest consecutive attention displacement in this comparison. This statistic describes the attention changes accompanying its prediction gains, complementing the benchmark metrics that evaluate relational grounding.

\paragraph{Attenuation rule.}
To examine whether attenuation benefits from prior inertia-selection frequency, we retain the full \ours token-selection rules, emergent-token reinforcement, and history updates, replacing only the inertia-token attenuation rule with a uniform coefficient. This coefficient is fixed before testing to the mean attenuation measured on held-out validation data. Table~\ref{tab:attenuation_rule_comparison} also reports the mean applied attenuation, $1-m_{t,j}$, over selected inertia tokens on the test set.

\begin{table}[t]
\centering
\caption{Comparison of attenuation rules under the full IVE selection protocol.}
\label{tab:attenuation_rule_comparison}
\footnotesize
\begin{tabularx}{\linewidth}{X*{4}{>{\centering\arraybackslash}X}}
\toprule
\textbf{Attenuation rule} & \textbf{Perc.}$\downarrow$ & \textbf{Cogn.}$\downarrow$ & $\mathbf{R_{score}}\uparrow$ & \textbf{Mean attenuation} \\
\midrule
Fixed, strength-matched & 35.21 & 31.60 & 66.81 & 0.061 \\
\rowcolor[HTML]{E8F2FE}
\textbf{History-frequency (IVE)} & \textbf{34.92} & \textbf{31.11} & \textbf{67.21} & 0.063 \\
\bottomrule
\end{tabularx}
\end{table}

History-frequency attenuation lowers both hallucination rates and improves $R_{score}$ over the fixed alternative (Table~\ref{tab:attenuation_rule_comparison}). Test-set mean attenuation is similar across conditions, with a slightly higher value for \ours. The comparison favors assigning attenuation according to prior inertia selections over applying one coefficient to every selected token under the evaluated protocol.

\paragraph{Raw and visual-normalized attention.}
We compare raw attention $a_{t,j}$ with visual-normalized attention $p_{t,j}$ to examine whether retaining total visual-attention strength benefits token selection. Each condition maintains an EMA in the corresponding input space and uses the fixed $K_+=8$, $K_-=32$ selection budgets and $1.10/0.90$ scaling factors from Table~\ref{tab:relevance_signal_comparison}. The raw-attention row in Table~\ref{tab:attention_input_comparison} reports the same matched-budget IVE-score condition.

\begin{table}[t]
\centering
\caption{Comparison of attention signals under matched selection budgets and scaling factors.}
\label{tab:attention_input_comparison}
\footnotesize
\begin{tabularx}{\linewidth}{X*{3}{>{\centering\arraybackslash}X}}
\toprule
\textbf{Score input} & \textbf{Perc.}$\downarrow$ & \textbf{Cogn.}$\downarrow$ & $\mathbf{R_{score}}\uparrow$ \\
\midrule
Vanilla (no intervention) & 37.67 & 33.99 & 64.39 \\
Visual-normalized attention ($p_{t,j}$) & 35.81 & 31.99 & 66.32 \\
\rowcolor[HTML]{E8F2FE}
\textbf{Raw attention ($a_{t,j}$; IVE score)} & \textbf{35.35} & \textbf{31.73} & \textbf{66.67} \\
\bottomrule
\end{tabularx}
\end{table}

Both inputs reduce hallucination rates relative to Vanilla, with raw attention providing a further improvement (Table~\ref{tab:attention_input_comparison}). The normalized-input result shows that the paired intervention remains effective when token rankings use only within-image allocation history. Under the tested scoring rule, retaining overall visual-attention strength alongside relative allocation yields an additional benefit with the same selection budgets and scaling factors.

\subsection{Matched Random Controls}
\label{sec:random_controls}

The preceding comparison varies token rankings under fixed budgets and scaling factors. We separately compare full \ours with random token perturbation at the same token budget and random visual-value modulation at the same magnitude. These controls examine whether unstructured perturbations produce gains comparable to history-conditioned selection and reweighting on Reefknot.

The two matching conditions address different aspects of the intervention. Matching the token budget keeps the number of affected tokens comparable, while matching the modulation magnitude controls the size of the value perturbation. They provide reference points for unstructured changes under comparable intervention budgets.

\begin{table}[t]
\centering
\caption{Matched random perturbation controls and \ours on Reefknot Y/N. The token control matches the token budget, and the value control matches the modulation magnitude.}
\label{tab:random_controls}
\footnotesize
\begin{tabularx}{\linewidth}{l*{3}{>{\centering\arraybackslash}X}}
\toprule
\textbf{Method} & $D_{\mathrm{attn}}$ & \textbf{Cogn.}$\downarrow$ & $R_{score}\uparrow$ \\
\midrule
Vanilla & 41.52 & 33.99 & 64.39 \\
+Random token perturbation (same budget) & 42.18 & 33.84 & 64.52 \\
+Random value modulation (same magnitude) & 42.63 & 33.71 & 64.60 \\
\rowcolor[HTML]{E8F2FE}
\textbf{+\ours} & \textbf{47.21} & \textbf{31.11} & \textbf{67.21} \\
\bottomrule
\end{tabularx}
\end{table}

Both random controls modestly improve $R_{score}$ over Vanilla, whereas \ours gives a larger improvement and a lower $Cogn.$ hallucination rate (Table~\ref{tab:random_controls}). The comparison favors structured selection and reweighting over the matched random perturbations.
Random perturbations also slightly increase consecutive attention displacement, but their prediction gains remain smaller than those of \ours.

\subsection{Additional Results on MME}
We report selected MME~\cite{fu2024mmecomprehensiveevaluationbenchmark} categories to examine whether the relation-oriented intervention preserves performance on other multimodal tasks. The comparison covers the same three backbones and reports category-specific scores.
The categories involve different demands on the visual input, including reading textual or numerical content and evaluating statements about that content. A category-level view shows how performance changes across these tasks, with each column expressed in its benchmark-specific score.

\begin{table}[t]
\centering
\caption{Scores on selected MME categories across three MLLMs. Com., Num., Text., and Code denote commonsense reasoning, numerical calculation, text translation, and code reasoning, respectively.}
\label{tab:mme}
\footnotesize
\begin{tabularx}{\linewidth}{ll*{4}{>{\centering\arraybackslash}X}}
\toprule
\textbf{Model} & \textbf{Method} & \textbf{Com.$\uparrow$} &
\textbf{Num.$\uparrow$} & \textbf{Text.$\uparrow$} & \textbf{Code.$\uparrow$} \\
\midrule
\multirow{10}{*}{LLaVA-1.5}
& Vanilla & 87.86 & 105.0 & 97.50 & 67.50 \\
& +VCD & 89.29 & 101.5 & 98.00 & 65.00 \\
& +OPERA & 92.14 & 107.0 & 92.50 & 58.00 \\
& +PAI & 91.43 & 106.5 & 91.50 & 56.50 \\
& +HELPD & 91.79 & 106.0 & 95.50 & 62.50 \\
& +DeCo & 90.71 & 102.0 & 100.0 & 70.00 \\
& +VASparse & 92.86 & 108.0 & 93.00 & 59.50 \\
& +MemVR & 93.57 & 109.0 & 101.5 & 70.50 \\
& +\textbf{IVE (Ours)} & \textbf{94.64} & \textbf{110.0} & \textbf{103.0} & \textbf{72.00} \\
& \cellcolor[HTML]{E8F2FE}\textbf{$\Delta$ vs. Vanilla}
& \cellcolor[HTML]{E8F2FE}\textcolor[HTML]{CB0000}{\textbf{+6.79}}
& \cellcolor[HTML]{E8F2FE}\textcolor[HTML]{CB0000}{\textbf{+5.00}}
& \cellcolor[HTML]{E8F2FE}\textcolor[HTML]{CB0000}{\textbf{+5.50}}
& \cellcolor[HTML]{E8F2FE}\textcolor[HTML]{CB0000}{\textbf{+4.50}} \\
\midrule
\multirow{10}{*}{InstructBLIP}
& Vanilla & 117.9 & 40.00 & 92.50 & 55.00 \\
& +VCD & 118.6 & 45.00 & 94.00 & 57.50 \\
& +OPERA & 121.4 & 47.50 & 89.50 & 52.00 \\
& +PAI & 120.7 & 47.00 & 89.00 & 51.50 \\
& +HELPD & 121.8 & 46.50 & 94.00 & 56.50 \\
& +DeCo & 120.4 & 46.00 & 96.00 & 59.50 \\
& +VASparse & 122.1 & 48.00 & 91.00 & 53.50 \\
& +MemVR & 122.9 & 47.50 & 97.00 & 60.00 \\
& +\textbf{IVE (Ours)} & \textbf{124.3} & \textbf{49.00} & \textbf{98.00} & \textbf{61.50} \\
& \cellcolor[HTML]{E8F2FE}\textbf{$\Delta$ vs. Vanilla}
& \cellcolor[HTML]{E8F2FE}\textcolor[HTML]{CB0000}{\textbf{+6.43}}
& \cellcolor[HTML]{E8F2FE}\textcolor[HTML]{CB0000}{\textbf{+9.00}}
& \cellcolor[HTML]{E8F2FE}\textcolor[HTML]{CB0000}{\textbf{+5.50}}
& \cellcolor[HTML]{E8F2FE}\textcolor[HTML]{CB0000}{\textbf{+6.50}} \\
\midrule
\multirow{10}{*}{Qwen2.5-VL-Instruct}
& Vanilla & 100.7 & 52.50 & 92.50 & 47.50 \\
& +VCD & 101.4 & 55.00 & 94.00 & 49.50 \\
& +OPERA & 103.6 & 56.00 & 89.50 & 46.50 \\
& +PAI & 102.9 & 55.50 & 89.00 & 46.00 \\
& +HELPD & 103.1 & 55.00 & 93.50 & 50.00 \\
& +DeCo & 102.3 & 54.50 & 96.00 & 52.50 \\
& +VASparse & 104.3 & 56.50 & 91.00 & 48.00 \\
& +MemVR & 104.6 & 57.00 & 97.00 & 53.00 \\
& +\textbf{IVE (Ours)} & \textbf{106.4} & \textbf{58.00} & \textbf{98.00} & \textbf{54.50} \\
& \cellcolor[HTML]{E8F2FE}\textbf{$\Delta$ vs. Vanilla}
& \cellcolor[HTML]{E8F2FE}\textcolor[HTML]{CB0000}{\textbf{+5.71}}
& \cellcolor[HTML]{E8F2FE}\textcolor[HTML]{CB0000}{\textbf{+5.50}}
& \cellcolor[HTML]{E8F2FE}\textcolor[HTML]{CB0000}{\textbf{+5.50}}
& \cellcolor[HTML]{E8F2FE}\textcolor[HTML]{CB0000}{\textbf{+7.00}} \\
\bottomrule
\end{tabularx}
\end{table}

\ours improves every selected category over Vanilla across all three backbones and remains competitive with the strongest baseline results. The gains extend across the reported tasks, supporting performance preservation beyond the primary relation-hallucination evaluations.

The baseline methods exhibit different profiles across the categories: some strengthen commonsense or numerical performance while falling below Vanilla on text or code, whereas \ours remains above Vanilla in all reported columns. The relation benchmarks provide the direct evaluation of the paper's primary objective, and these category-level results assess compatibility with other multimodal tasks.

\subsection{Radar Chart Results}
Figure~\ref{fig:radar} brings together the relation-focused results from Reefknot~\cite{zheng2025reefknot} and Hal-Eval~\cite{jiang2024hal} for LLaVA-1.5, InstructBLIP~\cite{dai2023instructblip}, and Qwen2.5-VL-Instruct~\cite{bai2025qwen2}. The five axes cover Reefknot Y/N Cogn. and $R_{score}$, Reefknot MCQ $R_{score}$, and Hal-Eval Relation accuracy and F1. \ours exhibits the strongest overall profile across the three backbones, complementing the detailed results in the main tables.

\subsection{Heatmap Comparison}
Figure~\ref{fig:heatmap} compares \ours with the amplification method PAI~\cite{liu2024paying}. In the illustrated response, \ours attends across several regions involved in the queried relation, whereas PAI remains concentrated within a smaller area. The contrast appears across successive output positions, providing a spatial view of the different attention patterns accompanying history-conditioned reweighting and amplification.

Reading the maps in temporal order reveals whether strongly attended regions continue to dominate across output positions. The displayed contrast captures the evolution of spatial emphasis and complements the distance summaries by retaining the image locations associated with each attention pattern.

\subsection{Object Hallucination on POPE}
POPE~\cite{Li-hallucination-2023} tests whether history-conditioned contribution reweighting also benefits object-existence judgments. Table~\ref{tab:pope} reports accuracy and F1 across the Random, Popular, and Adversarial settings for all three backbones. These settings vary the object queries, allowing the comparison to extend beyond one collection of candidate objects. Accuracy summarizes the binary decisions, while F1 reflects the balance between precision and recall for positive predictions. Parenthetical values report changes relative to Vanilla within the same backbone and setting.

\ours improves both metrics over Vanilla and the competing methods in each setting. Scores generally decrease from Random to Popular and Adversarial queries across the compared methods, while the advantage of \ours persists. This evaluation complements Reefknot by assessing temporal-relevance-based visual contribution updates on object-existence decisions.

\begin{table}[t]
\centering
\caption{
\textbf{Object-hallucination results on POPE.}
Accuracy and F1 (\%) across the Random, Popular, and Adversarial settings. Parenthetical values denote improvements over Vanilla.}
\label{tab:pope}
\newcommand{\popeval}[2]{#1\,{\scriptsize(#2)}}
\newcommand{\popebest}[2]{{\color[HTML]{CB0000}\textbf{#1\,{\scriptsize(#2)}}}}
\resizebox{\linewidth}{!}{%
\begin{tabular}{llcccccc}
\toprule
\multirow{2}{*}{\textbf{Category}} & \multirow{2}{*}{\textbf{Method}}
& \multicolumn{2}{c}{\textbf{LLaVA-1.5}}
& \multicolumn{2}{c}{\textbf{InstructBLIP}}
& \multicolumn{2}{c}{\textbf{Qwen2.5-VL-Instruct}} \\
\cmidrule(lr){3-4}\cmidrule(lr){5-6}\cmidrule(l){7-8}
& & \textbf{Acc.}$\uparrow$ & \textbf{F1}$\uparrow$
& \textbf{Acc.}$\uparrow$ & \textbf{F1}$\uparrow$
& \textbf{Acc.}$\uparrow$ & \textbf{F1}$\uparrow$ \\
\midrule
\multirow{8}{*}{Random}
& Vanilla & \popeval{83.29}{+0.00} & \popeval{81.33}{+0.00} & \popeval{80.71}{+0.00} & \popeval{80.41}{+0.00} & \popeval{85.27}{+0.00} & \popeval{85.30}{+0.00} \\
& +VCD & \popeval{86.27}{+2.98} & \popeval{86.34}{+5.01} & \popeval{84.53}{+3.82} & \popeval{83.68}{+3.27} & \popeval{88.60}{+3.33} & \popeval{88.67}{+3.37} \\
& +OPERA & \popeval{88.30}{+5.01} & \popeval{88.36}{+7.03} & \popeval{85.80}{+5.09} & \popeval{85.89}{+5.48} & \popeval{89.87}{+4.60} & \popeval{89.93}{+4.63} \\
& +DeCo & \popeval{88.13}{+4.84} & \popeval{88.20}{+6.87} & \popeval{84.87}{+4.16} & \popeval{84.79}{+4.38} & \popeval{89.63}{+4.36} & \popeval{89.68}{+4.38} \\
& +PAI & \popeval{88.42}{+5.13} & \popeval{88.51}{+7.18} & \popeval{85.20}{+4.49} & \popeval{85.29}{+4.88} & \popeval{89.77}{+4.50} & \popeval{89.82}{+4.52} \\
& +VASparse & \popeval{88.57}{+5.28} & \popeval{88.63}{+7.30} & \popeval{86.20}{+5.49} & \popeval{86.28}{+5.87} & \popeval{89.97}{+4.70} & \popeval{90.02}{+4.72} \\
& +MemVR & \popeval{88.83}{+5.54} & \popeval{88.89}{+7.56} & \popeval{86.57}{+5.86} & \popeval{86.64}{+6.23} & \popeval{90.40}{+5.13} & \popeval{90.45}{+5.15} \\
& \cellcolor[HTML]{E8F2FE}\textbf{+Ours} & \cellcolor[HTML]{E8F2FE}\popebest{90.53}{+7.24} & \cellcolor[HTML]{E8F2FE}\popebest{90.94}{+9.61} & \cellcolor[HTML]{E8F2FE}\popebest{90.20}{+9.49} & \cellcolor[HTML]{E8F2FE}\popebest{90.16}{+9.75} & \cellcolor[HTML]{E8F2FE}\popebest{92.10}{+6.83} & \cellcolor[HTML]{E8F2FE}\popebest{92.15}{+6.85} \\
\midrule
\multirow{8}{*}{Popular}
& Vanilla & \popeval{81.88}{+0.00} & \popeval{80.06}{+0.00} & \popeval{78.22}{+0.00} & \popeval{78.36}{+0.00} & \popeval{83.73}{+0.00} & \popeval{83.83}{+0.00} \\
& +VCD & \popeval{83.67}{+1.79} & \popeval{83.77}{+3.71} & \popeval{80.70}{+2.48} & \popeval{80.80}{+2.44} & \popeval{85.03}{+1.30} & \popeval{85.12}{+1.29} \\
& +OPERA & \popeval{84.60}{+2.72} & \popeval{84.69}{+4.63} & \popeval{81.97}{+3.75} & \popeval{82.07}{+3.71} & \popeval{86.33}{+2.60} & \popeval{86.41}{+2.58} \\
& +DeCo & \popeval{83.90}{+2.02} & \popeval{83.99}{+3.93} & \popeval{81.33}{+3.11} & \popeval{81.44}{+3.08} & \popeval{85.47}{+1.74} & \popeval{85.54}{+1.71} \\
& +PAI & \popeval{84.13}{+2.25} & \popeval{84.23}{+4.17} & \popeval{81.60}{+3.38} & \popeval{81.71}{+3.35} & \popeval{85.91}{+2.18} & \popeval{85.98}{+2.15} \\
& +VASparse & \popeval{84.77}{+2.89} & \popeval{84.84}{+4.78} & \popeval{82.17}{+3.95} & \popeval{82.27}{+3.91} & \popeval{86.57}{+2.84} & \popeval{86.64}{+2.81} \\
& +MemVR & \popeval{85.00}{+3.12} & \popeval{85.09}{+5.03} & \popeval{82.50}{+4.28} & \popeval{82.60}{+4.24} & \popeval{86.83}{+3.10} & \popeval{86.91}{+3.08} \\
& \cellcolor[HTML]{E8F2FE}\textbf{+Ours} & \cellcolor[HTML]{E8F2FE}\popebest{86.90}{+5.02} & \cellcolor[HTML]{E8F2FE}\popebest{87.01}{+6.95} & \cellcolor[HTML]{E8F2FE}\popebest{84.73}{+6.51} & \cellcolor[HTML]{E8F2FE}\popebest{85.45}{+7.09} & \cellcolor[HTML]{E8F2FE}\popebest{89.70}{+5.97} & \cellcolor[HTML]{E8F2FE}\popebest{89.75}{+5.92} \\
\midrule
\multirow{8}{*}{Adversarial}
& Vanilla & \popeval{76.47}{+0.00} & \popeval{76.61}{+0.00} & \popeval{74.37}{+0.00} & \popeval{74.51}{+0.00} & \popeval{81.87}{+0.00} & \popeval{81.98}{+0.00} \\
& +VCD & \popeval{77.20}{+0.73} & \popeval{77.34}{+0.73} & \popeval{76.80}{+2.43} & \popeval{76.94}{+2.43} & \popeval{83.47}{+1.60} & \popeval{83.57}{+1.59} \\
& +OPERA & \popeval{78.00}{+1.53} & \popeval{78.13}{+1.52} & \popeval{77.50}{+3.13} & \popeval{77.63}{+3.12} & \popeval{84.67}{+2.80} & \popeval{84.76}{+2.78} \\
& +DeCo & \popeval{77.57}{+1.10} & \popeval{77.68}{+1.07} & \popeval{77.07}{+2.70} & \popeval{77.21}{+2.70} & \popeval{83.87}{+2.00} & \popeval{83.97}{+1.99} \\
& +PAI & \popeval{77.83}{+1.36} & \popeval{77.96}{+1.35} & \popeval{77.37}{+3.00} & \popeval{77.49}{+2.98} & \popeval{84.30}{+2.43} & \popeval{84.39}{+2.41} \\
& +VASparse & \popeval{78.30}{+1.83} & \popeval{78.42}{+1.81} & \popeval{78.13}{+3.76} & \popeval{78.53}{+4.02} & \popeval{84.87}{+3.00} & \popeval{84.93}{+2.95} \\
& +MemVR & \popeval{78.60}{+2.13} & \popeval{78.73}{+2.12} & \popeval{78.40}{+4.03} & \popeval{78.42}{+3.91} & \popeval{85.47}{+3.60} & \popeval{85.56}{+3.58} \\
& \cellcolor[HTML]{E8F2FE}\textbf{+Ours} & \cellcolor[HTML]{E8F2FE}\popebest{80.53}{+4.06} & \cellcolor[HTML]{E8F2FE}\popebest{82.17}{+5.56} & \cellcolor[HTML]{E8F2FE}\popebest{80.50}{+6.13} & \cellcolor[HTML]{E8F2FE}\popebest{81.70}{+7.19} & \cellcolor[HTML]{E8F2FE}\popebest{88.17}{+6.30} & \cellcolor[HTML]{E8F2FE}\popebest{88.23}{+6.25} \\
\bottomrule
\end{tabular}
}
\end{table}

\subsection{Object Hallucination in Generated Captions}
\label{sec:chair_results}

Object-existence questions constrain the model to a short response, while captioning allows it to introduce objects throughout generation. We therefore report CHAIR~\cite{Rohrbach2018ObjectHI} results on a COCO subset. $\mathrm{CHAIR}_i$ measures the fraction of mentioned objects that are hallucinated, and $\mathrm{CHAIR}_s$ measures the fraction of captions containing at least one hallucinated object. We report both overall measures and stratify $\mathrm{CHAIR}_i$ by scene density and response length.

The density and length groups provide two complementary views of the captioning results. Density groups distinguish scenes by object count, and length groups distinguish generated responses by token count. These marginal stratifications show how each method behaves in more crowded scenes and longer responses.

\begin{table}[t]
\centering
\caption{CHAIR results (\%) overall and by scene density and response length. Density counts objects, and length counts output tokens. Lower values are better.}
\label{tab:chair_stratified}
\resizebox{\linewidth}{!}{%
\begin{tabular}{lcccccccc}
\toprule
\multirow{3}{*}{\textbf{Method}} &
\multicolumn{2}{c}{\textbf{Overall}} &
\multicolumn{3}{c}{\textbf{Scene density} ($\mathrm{CHAIR}_i$)} &
\multicolumn{3}{c}{\textbf{Response length} ($\mathrm{CHAIR}_i$)} \\
\cmidrule(lr){2-3}\cmidrule(lr){4-6}\cmidrule(lr){7-9}
& \multirow{2}{*}{$\mathrm{CHAIR}_i\downarrow$}
& \multirow{2}{*}{$\mathrm{CHAIR}_s\downarrow$}
& Sparse & Medium & Dense & Short & Medium & Long \\
& & & $\leq5$ & $6$--$10$ & $\geq11$ & $\leq50$ & $51$--$100$ & $>100$ \\
\midrule
Vanilla & 52.4 & 68.1 & 38.2 & 54.7 & 71.3 & 44.3 & 53.8 & 61.9 \\
+PAI & 51.8 & 67.5 & 37.9 & 54.1 & 71.6 & 44.1 & 53.4 & 62.7 \\
+MemVR & 51.2 & 66.9 & 37.5 & 53.8 & 70.9 & 43.7 & 52.9 & 61.4 \\
\rowcolor[HTML]{E8F2FE}
\textbf{+\ours} & \textbf{50.3} & \textbf{65.8} & \textbf{37.1} & \textbf{53.2} & \textbf{70.8} & \textbf{43.2} & \textbf{52.4} & \textbf{61.1} \\
\bottomrule
\end{tabular}
}
\end{table}

Hallucination rates rise with scene density and response length for all compared methods (Table~\ref{tab:chair_stratified}). \ours yields the lowest overall rates and remains below the baselines in each density and length group, including dense scenes and long responses. These results extend the object-hallucination comparison to the unsupported object mentions introduced during caption generation.

\ours maintains lower hallucination rates in the densest scenes and longest responses, with smaller gains over Vanilla than in the other groups. The benefit concerns the reliability of the object content that is generated, complementing the object-existence evaluation on POPE.

\subsection{Full Configuration Sensitivity}
Table~\ref{tab:coefficient} reports coefficient sensitivity across the three backbones on Reefknot and POPE. Each sweep varies one coefficient around the validation-selected configuration while keeping the remaining settings fixed. The attenuation coefficient $\alpha$ bounds the reduction of persistent contributions at each update, $\gamma$ controls smoothing of the recent attention history, and $\tau$ sets the threshold for identifying emergent tokens. The main text retains the LLaVA-1.5 subset, while the full table shows the corresponding local variations on the other backbones.

These coefficients influence different parts of the update. Changing $\alpha$ adjusts attenuation after an inertia token has been selected. Changing $\gamma$ modifies the historical reference against which current attention is compared, with larger values assigning more weight to the older reference. Changing $\tau$ alters the threshold for the emergent-token branch. Examining them separately helps identify whether performance is sensitive to the magnitude of reweighting, the construction of attention history, or the boundary used for token selection.

\begin{table}[t]
\centering
\caption{Full coefficient sensitivity of \ours across three backbones on Reefknot and POPE.}
\label{tab:coefficient}
\footnotesize
\begin{tabularx}{\linewidth}{lll*{4}{>{\centering\arraybackslash}X}}
\toprule
\multirow{2}{*}{\textbf{Model}} & \multirow{2}{*}{\textbf{Coeff.}}
& \multirow{2}{*}{\textbf{Val.}}
& \multicolumn{2}{c}{\textbf{Reefknot}}
& \multicolumn{2}{c}{\textbf{POPE}} \\
\cmidrule(lr){4-5}\cmidrule(l){6-7}
& & & \textbf{Perc.}$\downarrow$ & \textbf{Cogn.}$\downarrow$
& \textbf{Acc.}$\uparrow$ & \textbf{F1}$\uparrow$ \\
\midrule
\multirow{9}{*}{LLaVA-1.5}
& \multirow{3}{*}{$\alpha$} & 0.05 & 36.84 & 32.72 & 89.20 & 89.27 \\
& & \cellcolor[HTML]{E8F2FE}0.10 & \cellcolor[HTML]{E8F2FE}34.92 & \cellcolor[HTML]{E8F2FE}31.11 & \cellcolor[HTML]{E8F2FE}90.53 & \cellcolor[HTML]{E8F2FE}90.94 \\
& & 0.15 & 36.60 & 32.43 & 89.57 & 89.62 \\
\cmidrule(lr){2-7}
& \multirow{3}{*}{$\gamma$} & 0.08 & 36.56 & 32.48 & 89.77 & 89.82 \\
& & \cellcolor[HTML]{E8F2FE}0.09 & \cellcolor[HTML]{E8F2FE}34.92 & \cellcolor[HTML]{E8F2FE}31.11 & \cellcolor[HTML]{E8F2FE}90.53 & \cellcolor[HTML]{E8F2FE}90.94 \\
& & 0.10 & 35.88 & 32.07 & 90.53 & 90.94 \\
\cmidrule(lr){2-7}
& \multirow{3}{*}{$\tau$} & 2.95 & 36.72 & 32.44 & 89.63 & 89.68 \\
& & \cellcolor[HTML]{E8F2FE}3.00 & \cellcolor[HTML]{E8F2FE}34.92 & \cellcolor[HTML]{E8F2FE}31.11 & \cellcolor[HTML]{E8F2FE}90.53 & \cellcolor[HTML]{E8F2FE}90.94 \\
& & 3.05 & 35.93 & 32.10 & 90.27 & 90.32 \\
\midrule
\multirow{9}{*}{InstructBLIP}
& \multirow{3}{*}{$\alpha$} & 0.05 & 44.28 & 41.73 & 87.23 & 87.31 \\
& & \cellcolor[HTML]{E8F2FE}0.10 & \cellcolor[HTML]{E8F2FE}42.94 & \cellcolor[HTML]{E8F2FE}39.31 & \cellcolor[HTML]{E8F2FE}90.20 & \cellcolor[HTML]{E8F2FE}90.16 \\
& & 0.15 & 43.88 & 41.34 & 88.93 & 88.95 \\
\cmidrule(lr){2-7}
& \multirow{3}{*}{$\gamma$} & 0.09 & 44.16 & 41.57 & 89.97 & 90.02 \\
& & 0.10 & 43.70 & 40.79 & 90.07 & 90.12 \\
& & \cellcolor[HTML]{E8F2FE}0.11 & \cellcolor[HTML]{E8F2FE}42.94 & \cellcolor[HTML]{E8F2FE}39.31 & \cellcolor[HTML]{E8F2FE}90.20 & \cellcolor[HTML]{E8F2FE}90.16 \\
\cmidrule(lr){2-7}
& \multirow{3}{*}{$\tau$} & 3.00 & 43.75 & 40.60 & 90.20 & 90.16 \\
& & \cellcolor[HTML]{E8F2FE}3.05 & \cellcolor[HTML]{E8F2FE}42.94 & \cellcolor[HTML]{E8F2FE}39.31 & \cellcolor[HTML]{E8F2FE}90.20 & \cellcolor[HTML]{E8F2FE}90.16 \\
& & 3.10 & 43.88 & 40.99 & 90.10 & 90.11 \\
\midrule
\multirow{9}{*}{Qwen2.5-VL-Instruct}
& \multirow{3}{*}{$\alpha$} & 0.10 & 36.18 & 19.22 & 91.80 & 91.85 \\
& & \cellcolor[HTML]{E8F2FE}0.15 & \cellcolor[HTML]{E8F2FE}34.91 & \cellcolor[HTML]{E8F2FE}17.60 & \cellcolor[HTML]{E8F2FE}92.10 & \cellcolor[HTML]{E8F2FE}92.15 \\
& & 0.20 & 36.11 & 19.08 & 91.93 & 91.98 \\
\cmidrule(lr){2-7}
& \multirow{3}{*}{$\gamma$} & 0.11 & 36.12 & 19.10 & 91.73 & 91.78 \\
& & \cellcolor[HTML]{E8F2FE}0.12 & \cellcolor[HTML]{E8F2FE}34.91 & \cellcolor[HTML]{E8F2FE}17.60 & \cellcolor[HTML]{E8F2FE}92.10 & \cellcolor[HTML]{E8F2FE}92.15 \\
& & 0.13 & 36.20 & 19.18 & 91.68 & 91.74 \\
\cmidrule(lr){2-7}
& \multirow{3}{*}{$\tau$} & 3.05 & 36.08 & 19.01 & 91.88 & 91.94 \\
& & \cellcolor[HTML]{E8F2FE}3.10 & \cellcolor[HTML]{E8F2FE}34.91 & \cellcolor[HTML]{E8F2FE}17.60 & \cellcolor[HTML]{E8F2FE}92.10 & \cellcolor[HTML]{E8F2FE}92.15 \\
& & 3.15 & 36.14 & 19.12 & 91.79 & 91.84 \\
\bottomrule
\end{tabularx}
\end{table}

Every tested setting improves over Vanilla on both Reefknot and POPE, while the selected settings yield the lowest $Cogn.$ rates within each backbone's sweeps. Gains persist throughout the tested neighborhoods, with their magnitude varying across configurations. This analysis examines neighborhoods around the tuned settings, complementing the shared-configuration comparison in the main text.

The relation-oriented metric varies non-monotonically across each coefficient sweep. The selected settings balance the effects of the relevance reference and contribution adjustment within the tested neighborhoods. The POPE columns show how the same local changes affect object-level evaluation. The unified-configuration experiment applies one shared choice across backbones, complementing these model-specific local sweeps.

\subsection{Ablation on Token Selection Layers}
The selection layer determines where \ours reads the attention signal, while value recalibration continues to apply across transformer layers. Table~\ref{tab:layer_ablation} varies the selection layer while keeping the remaining method fixed. The chosen mid-to-late layers yield the lowest Reefknot~\cite{zheng2025reefknot} hallucination rates for each backbone and retain competitive accuracy and F1 on POPE~\cite{Li-hallucination-2023}. POPE F1 can peak at a neighboring layer. Earlier layers also improve over Vanilla, with smaller gains on Reefknot.

Changing the selection layer changes both the attention history and the point at which recalibration enters the current forward pass. Cache updates still span all layers. The results favor the chosen layers for relation evaluation, with gains also retained at neighboring layers.

\begin{table}[t]
\centering
\caption{Ablation study on token selection layers of \ours.}
\label{tab:layer_ablation}
\footnotesize
\begin{tabularx}{\linewidth}{ll*{4}{>{\centering\arraybackslash}X}}
\toprule
\multirow{2}{*}{\textbf{Model}} & \multirow{2}{*}{\textbf{Layer}}
& \multicolumn{2}{c}{\textbf{Reefknot}}
& \multicolumn{2}{c}{\textbf{POPE}} \\
\cmidrule(lr){3-4}\cmidrule(l){5-6}
& & \textbf{Perc.}$\downarrow$ & \textbf{Cogn.}$\downarrow$
& \textbf{Acc.}$\uparrow$ & \textbf{F1}$\uparrow$ \\
\midrule
\multirow{3}{*}{LLaVA-1.5}
& 16 & 36.14 & 32.36 & 90.18 & 90.24 \\
& 18 & 35.96 & 32.18 & 90.41 & 90.48 \\
& \cellcolor[HTML]{E8F2FE}20
& \cellcolor[HTML]{E8F2FE}\textbf{34.92}
& \cellcolor[HTML]{E8F2FE}\textbf{31.11}
& \cellcolor[HTML]{E8F2FE}\textbf{90.53}
& \cellcolor[HTML]{E8F2FE}\textbf{90.94} \\
\midrule
\multirow{3}{*}{InstructBLIP}
& 16 & 44.02 & 40.97 & 89.94 & 90.01 \\
& 18 & 43.84 & 40.71 & 90.18 & 90.26 \\
& \cellcolor[HTML]{E8F2FE}20
& \cellcolor[HTML]{E8F2FE}\textbf{42.94}
& \cellcolor[HTML]{E8F2FE}\textbf{39.31}
& \cellcolor[HTML]{E8F2FE}\textbf{90.20}
& \cellcolor[HTML]{E8F2FE}\textbf{90.16} \\
\midrule
\multirow{3}{*}{Qwen2.5-VL-Instruct}
& 14 & 36.09 & 19.14 & 91.72 & 91.76 \\
& 16 & 35.94 & 18.95 & 91.97 & 92.02 \\
& \cellcolor[HTML]{E8F2FE}18
& \cellcolor[HTML]{E8F2FE}\textbf{34.91}
& \cellcolor[HTML]{E8F2FE}\textbf{17.60}
& \cellcolor[HTML]{E8F2FE}\textbf{92.10}
& \cellcolor[HTML]{E8F2FE}\textbf{92.15} \\
\bottomrule
\end{tabularx}
\end{table}

\begin{figure}[t]
    \centering
    \includegraphics[width=0.78\linewidth,trim=8.30bp 0 632.20bp 0,clip]{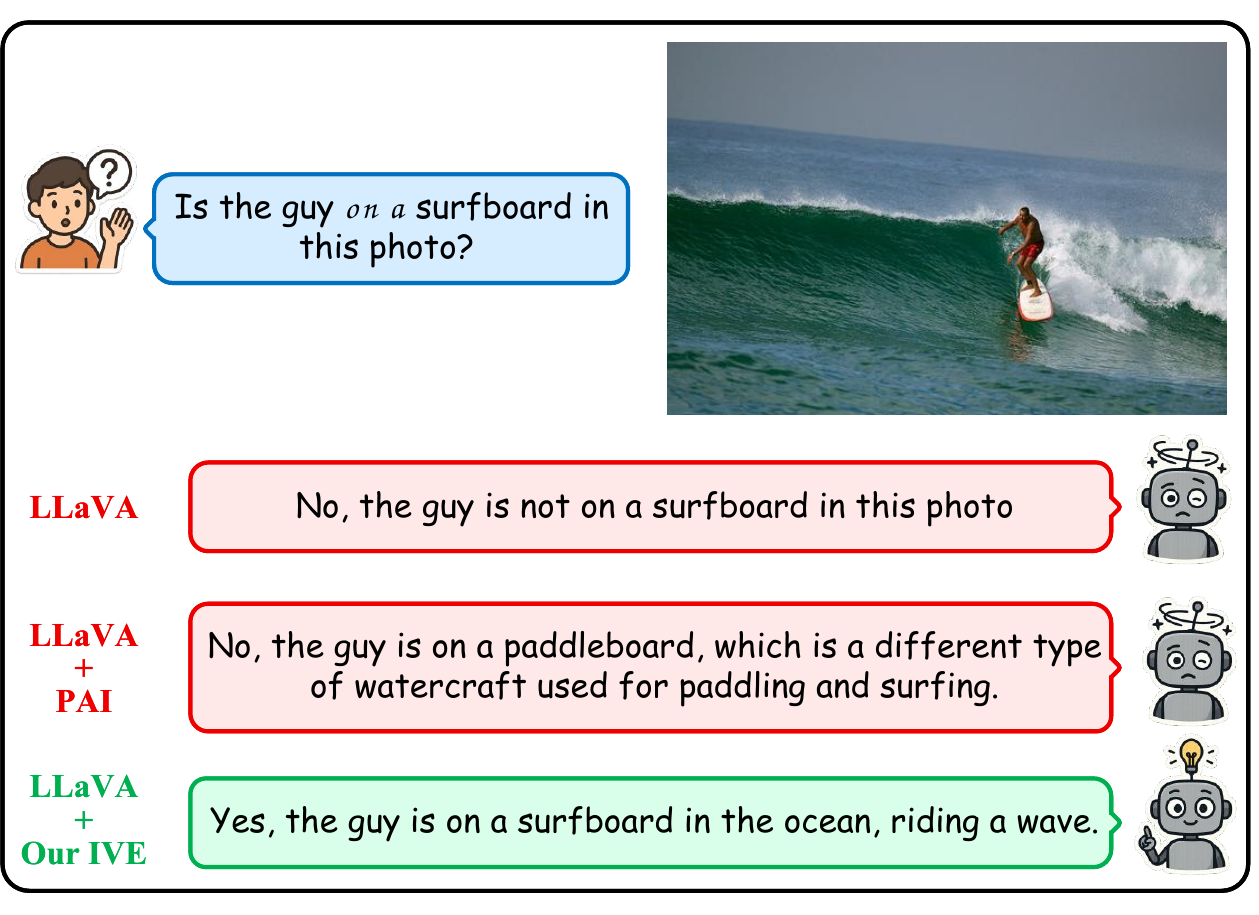}
    \par\vspace{1mm}
    \includegraphics[width=0.78\linewidth,trim=637.15bp 0 3.35bp 0,clip]{fig/case1.pdf}
    \caption{Case 1.}
    \label{fig:case_reef1}
\end{figure}

\section{Case Studies}
\label{sec:case_studies_appendix}
\begin{figure}[t]
    \centering
    \includegraphics[width=0.8\linewidth]{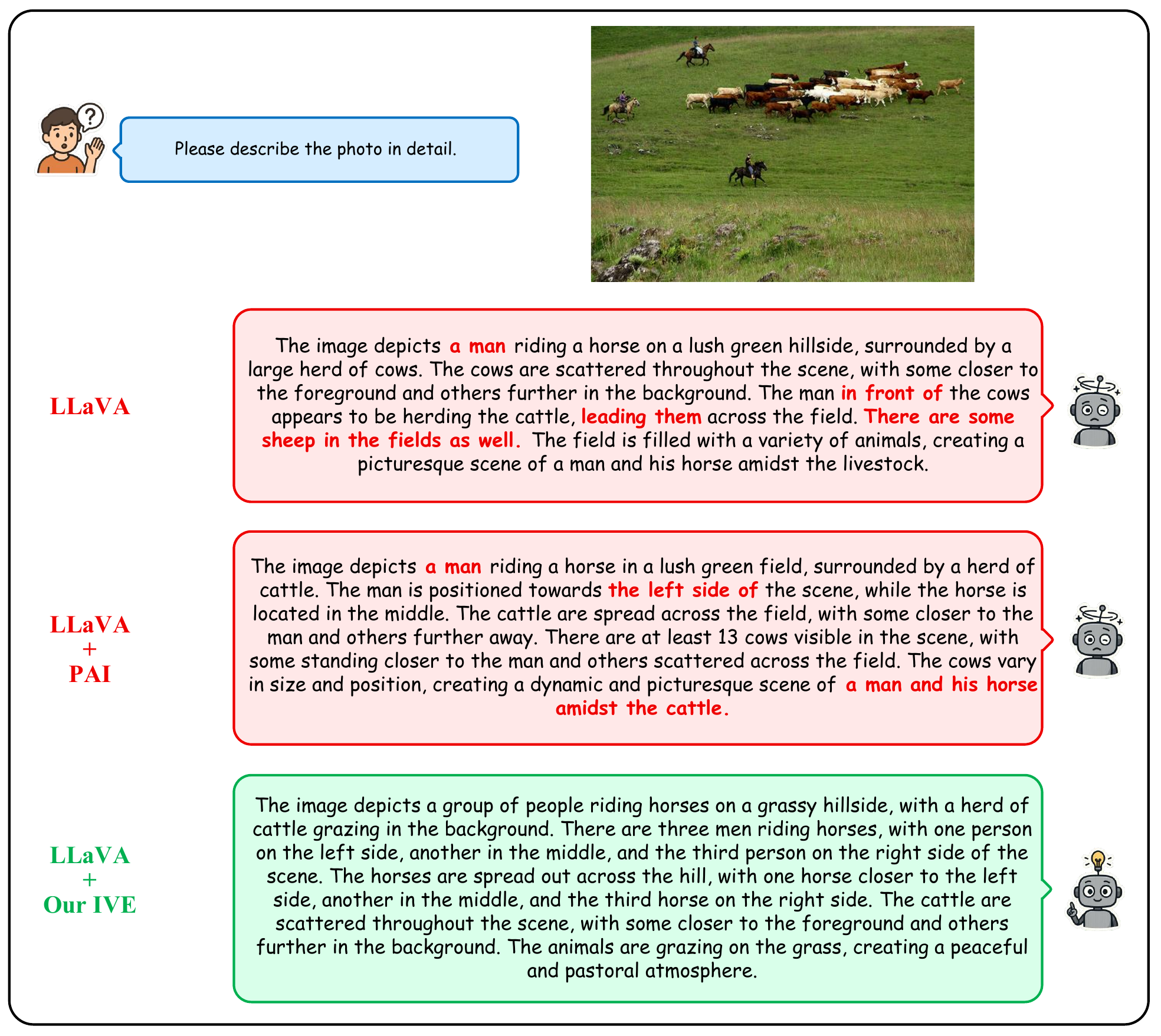}
    \caption{Case 2.}
    \label{fig:case_reef2}
\end{figure}

\begin{figure}[t]
    \centering
    \includegraphics[width=0.8\linewidth]{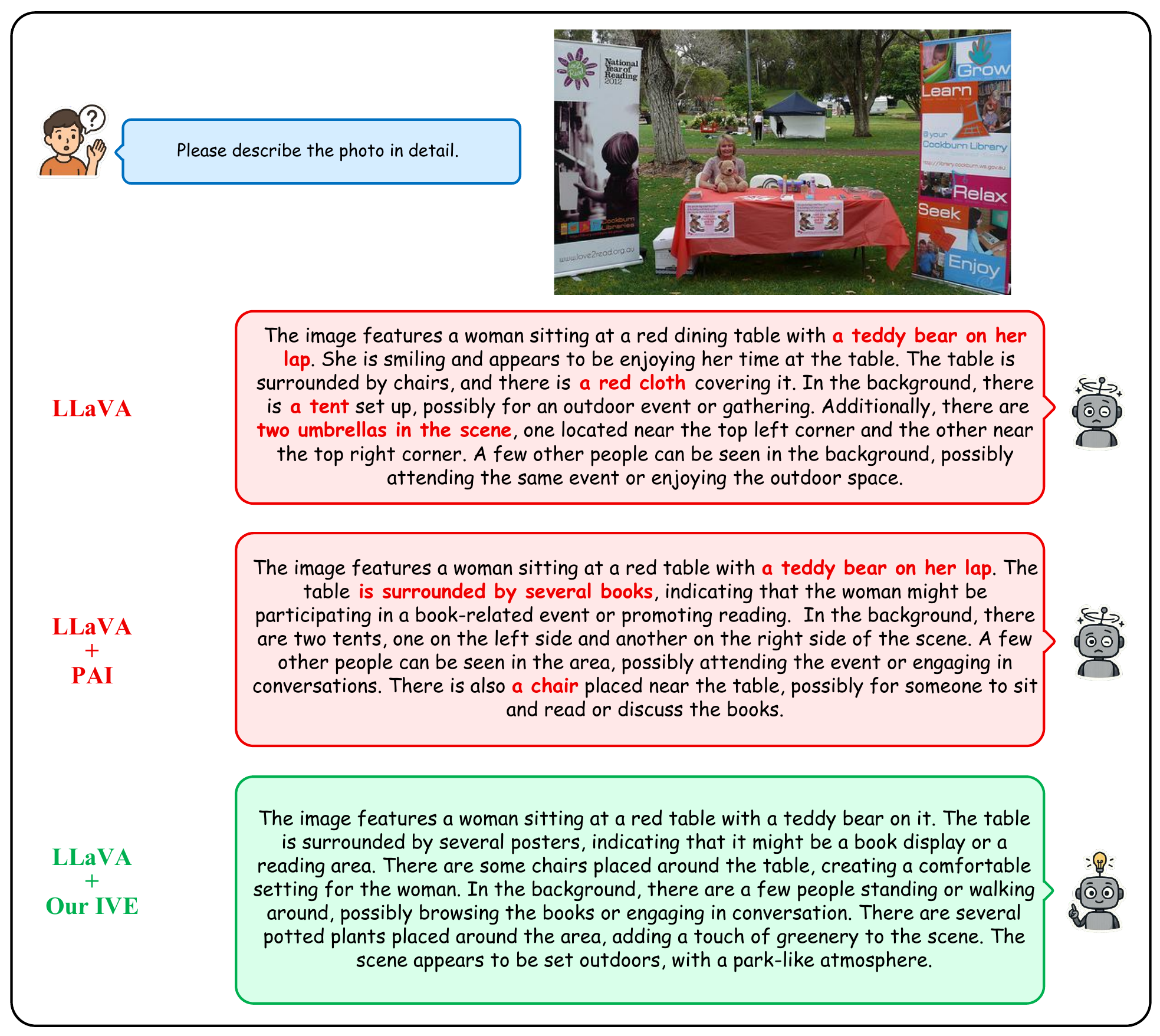}
    \caption{Case 3.}
    \label{fig:case_reef3}
\end{figure}

\begin{figure}[t]
    \centering
    \includegraphics[width=0.8\linewidth]{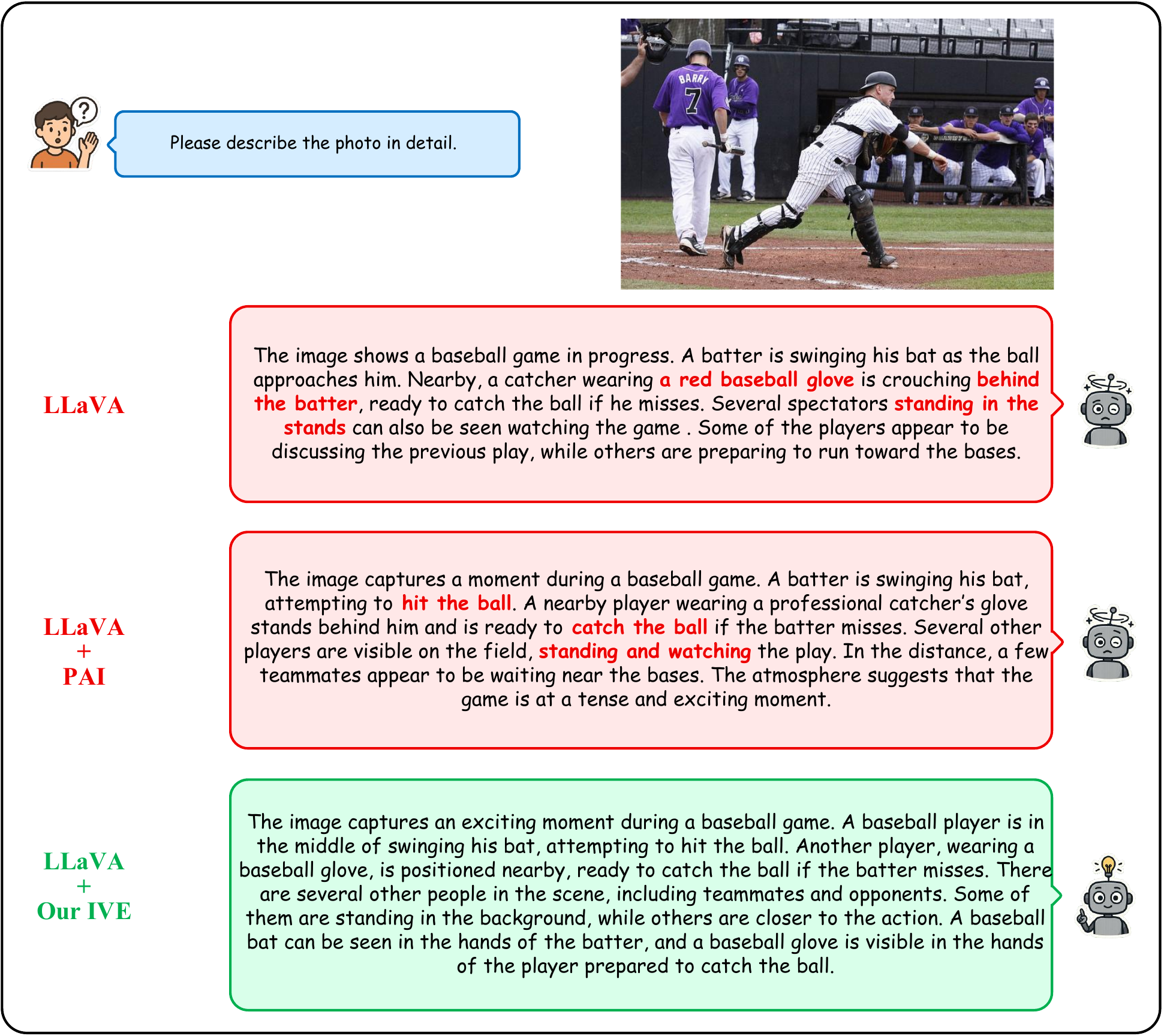}
    \caption{Case 4.}
    \label{fig:case_reef7}
\end{figure}

\begin{figure}[t]
    \centering
    \includegraphics[width=0.8\linewidth]{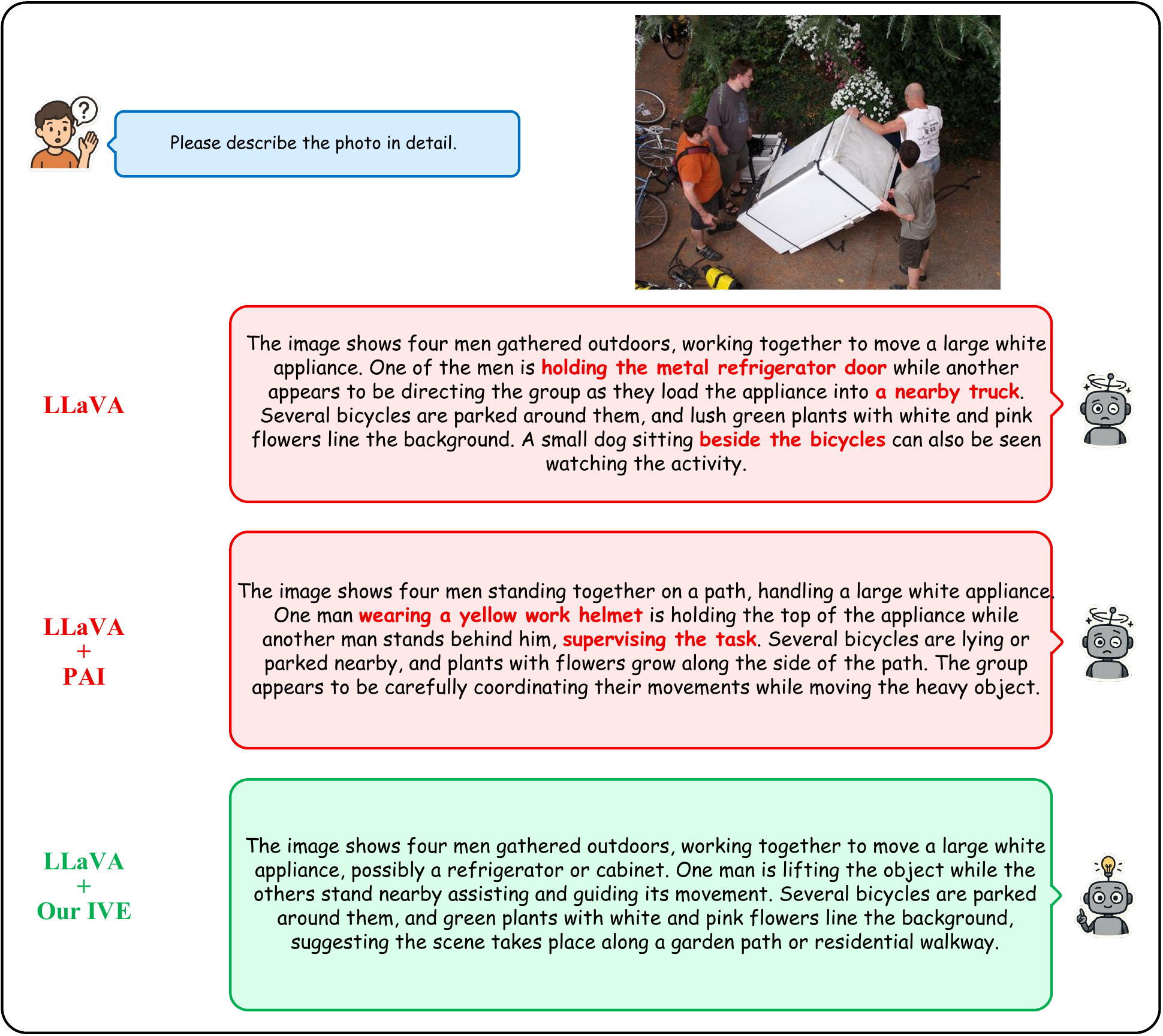}
    \caption{Case 5.}
    \label{fig:case_reef8}
\end{figure}

\begin{figure}[t]
    \centering
    \includegraphics[width=0.78\linewidth,trim=7.28bp 0 630.62bp 0,clip]{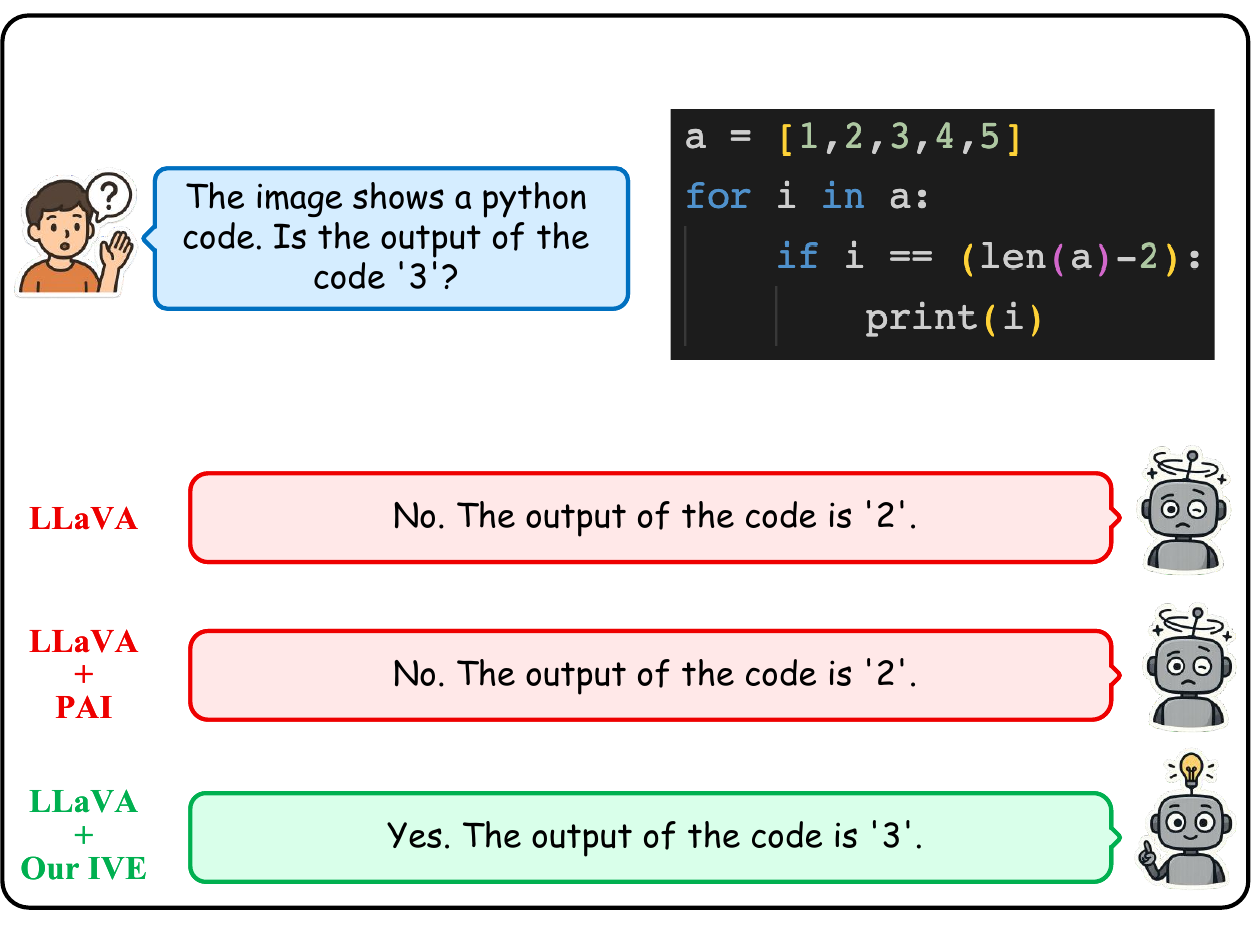}
    \par\vspace{1mm}
    \includegraphics[width=0.78\linewidth,trim=636.13bp 0 1.77bp 0,clip]{fig/case4.pdf}
    \caption{Case 6.}
    \label{fig:case_mme}
\end{figure}

\begin{figure}[t]
    \centering
    \includegraphics[width=0.78\linewidth,trim=4.69bp 0 629.02bp 0,clip]{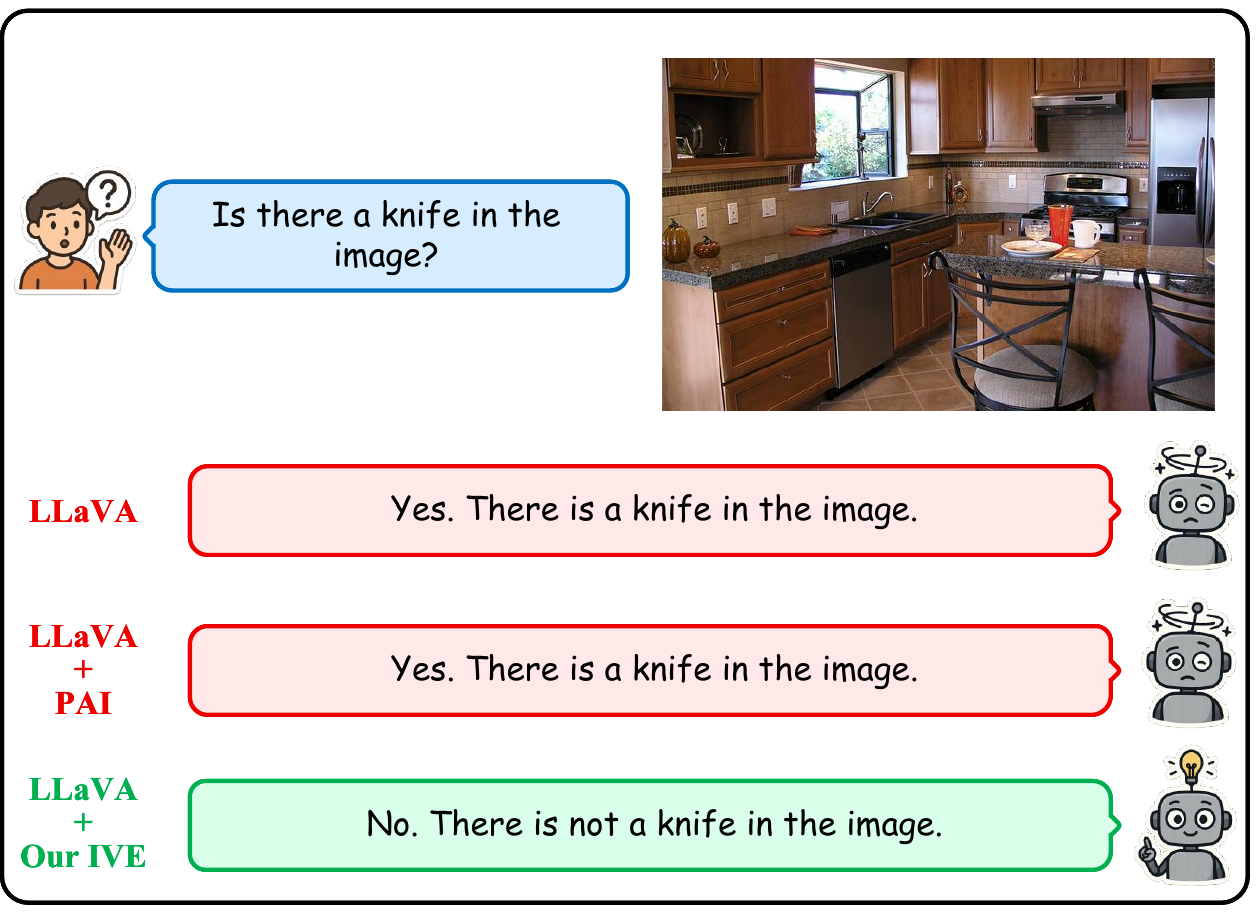}
    \par\vspace{1mm}
    \includegraphics[width=0.78\linewidth,trim=633.54bp 0 0.17bp 0,clip]{fig/case5.pdf}
    \caption{Case 7.}
    \label{fig:case_pope}
\end{figure}

\begin{figure}[t]
    \centering
    \includegraphics[width=0.78\linewidth,trim=6.38bp 0 629.13bp 0,clip]{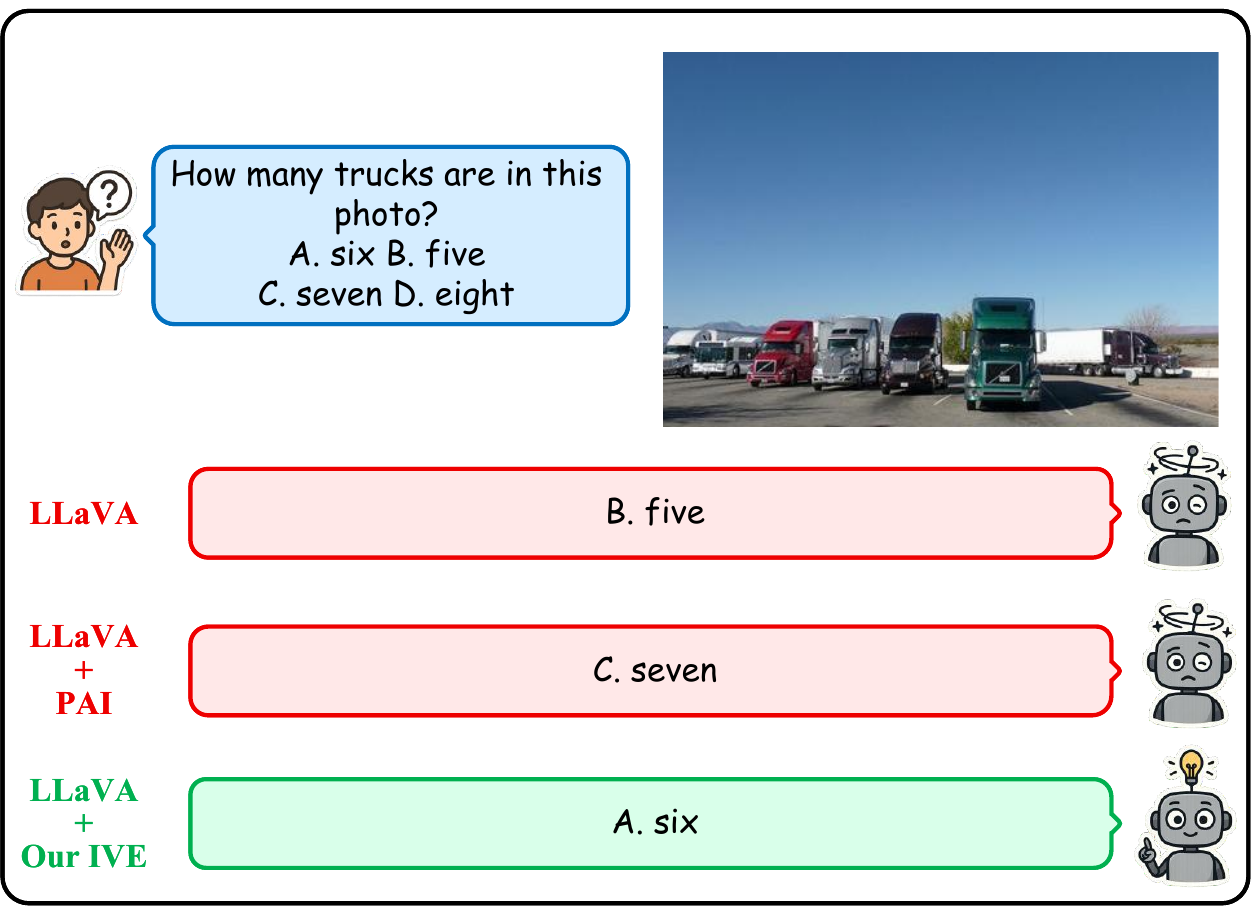}
    \par\vspace{1mm}
    \includegraphics[width=0.78\linewidth,trim=635.23bp 0 0.28bp 0,clip]{fig/case6.pdf}
    \caption{Case 8.}
    \label{fig:case_mmbench}
\end{figure}

Figures~\ref{fig:case_reef1}--\ref{fig:case_mmbench} present Cases 1--8 comparing Vanilla and \ours. Cases 1--5 are drawn from Reefknot~\cite{zheng2025reefknot}, Case 6 from MME~\cite{fu2024mmecomprehensiveevaluationbenchmark}, Case 7 from POPE~\cite{Li-hallucination-2023}, and Case 8 from MMBench~\cite{liu2024mmbench}. Each comparison uses LLaVA-1.5~\cite{liu2024improved} with the same visual input and prompt.

Cases 1--5 illustrate unsupported relations predicted by the baseline, showing how responses can mention the correct entities yet misrepresent their relationship. The paired responses highlight differences in the relation-bearing assertions, complementing the aggregate relation-hallucination results.

Case 7 concerns whether a named object is present, contrasting the methods' object-existence decisions. Cases 6 and 8 illustrate differences in responses to the task-specific questions of the broader capability evaluations.

\end{document}